%% file: iclr2021_conference.tex
\documentclass{article} 
\usepackage{iclr2021_conference,times}

\input{math_commands.tex}

\usepackage{graphicx}
\usepackage{wrapfig}
\usepackage{caption}
\usepackage{lipsum,booktabs}
\usepackage{subfiles}
\usepackage{import}
\usepackage{longtable}
\usepackage{tabularx} 
\usepackage{changepage}
\usepackage{float}
\usepackage[compact]{titlesec}  
\usepackage{siunitx}
\usepackage{makecell}
\usepackage{url}
\usepackage{natbib,subfig}
\usepackage[inline]{enumitem}

\newcommand*\samethanks[1][\value{footnote}]{\footnotemark[#1]}

\title{\centering{GeDi: Generative Discriminator guided Sequence Generation}}

\author{Ben Krause\thanks{\,Equal Contribution}, \,Akhilesh Deepak Gotmare\samethanks, \,Bryan McCann\thanks{Work performed while at Salesforce Research}, \,Nitish Shirish Keskar \\
 \textbf{Shafiq Joty, Richard Socher\samethanks, Nazneen Fatema Rajani} \\
 \\
 Salesforce Research\\
\texttt{\{bkrause,akhilesh.gotmare\}@salesforce.com} \\
}

\iclrfinalcopy 
\begin{document}

\maketitle
\begin{abstract}

While large-scale language models (LMs) are able to imitate the distribution of natural language well enough to generate realistic text, it is difficult to control which regions of the distribution they generate. This is especially problematic because datasets used for training large LMs usually contain significant toxicity, hate, bias, and negativity. We propose GeDi as an efficient method for using smaller LMs as generative discriminators to guide generation from large LMs to make them safer and more controllable. GeDi guides generation at each step by computing classification probabilities for all possible next tokens via Bayes rule by normalizing over two class-conditional distributions; one conditioned on the desired attribute, or \emph{control code}, and another conditioned on the undesired attribute, or \emph{anti control code}. We find that GeDi gives stronger controllability than the state of the art method while also achieving generation speeds more than $30$ times faster. Additionally, training GeDi on only four topics allows us to controllably generate new topics zero-shot from just a keyword, unlocking a new capability that previous controllable generation methods do not have. Lastly, we show that GeDi can make GPT-2 (1.5B parameters) significantly less toxic without sacrificing linguistic quality, making it by far the most practical existing method for detoxifying large language models while maintaining a fast generation speed.\footnote{\scriptsize Code available at \url{https://github.com/salesforce/GeDi}, includes GeDi-guided GPT-3 generation using OpenAI API. }

\end{abstract}

\section{Introduction}

Natural language generation has seen great progress with the advent of Transformers \citep{vaswani2017attention} and large scale training \citep{radford2017learning,radford2018improving,radford2019language, brown2020language}. Large language models (LMs) like GPT-2 \citep{radford2019language} and GPT-3 \citep{brown2020language} are able to learn the distribution of their training set well enough to generate realistic text. However, simply imitating the distribution of the training data during generation has many drawbacks; large-scale text training sets are crawled from the web which is imbued with toxicity, bias, hate, and misinformation. Methods for better controlling or filtering generation are valuable for making LMs trained on such data safer and more generally useful for downstream applications.

Existing approaches to controlling LMs have limitations. Class-conditional LMs (CC-LMs) such as CTRL \citep{keskar2019ctrl} attempt to control text generation by conditioning on a \emph{control code}, which is an attribute variable representing a data source. However, CTRL is not as useful for controlling what \emph{not}  to generate (i.e. toxicity). Furthermore, using a specific control code can reduce sample diversity across prompts, as samples will generally resemble the data source of the control code. Another approach is to use discriminators to steer generation, but existing methods to do this are very computationally intensive. Weighted decoding \citep{holtzman2018learning} requires feeding candidate next tokens into a discriminator, and thus scales linearly in computation with the number of tokens to be re-weighted. Plug and Play LM \citep[PPLM]{dathathri2020plug} applies up to 10 updates to the generating LM's latent states per time step using gradients from a discriminator, also making it many times slower than generating from the LM directly.

We present GeDi\footnote{\scriptsize pronounced ``Jedi''} as an algorithm for efficiently guiding generation from large LMs to make them safer and more controllable. Our proposed method uses CC-LMs as generative discriminators (GeDis) to guide language generation towards desired attributes. The methods we develop include:

\begin{itemize}
    \item GeDi-guided contrastive generation: We show how CC-LMs can be used as generative discriminators to compute classification likelihoods for all candidate next tokens during generation using Bayes rule, saving many thousand-fold in computation as compared with using a standard (non-generative) discriminator to compute this for large vocabulary sizes. We then show how these likelihoods can guide generation from large language models via weighted decoding and filtering [Section \ref{sec:contrastive}].  
    \item GeDi training: We train CC-LMs with a hybrid generative-discriminative loss to make them better classifiers, making them more powerful discriminators for GeDi-guided contrastive generation [Section \ref{sec:training}]. 

\end{itemize}

Our experimental results verify the ability of GeDi to control generation in a variety of settings while maintaining linguistic quality on par with strong language models. We apply GeDi (345M parameters) to guide generation from the GPT2-XL model (1.5B parameters), and find that:
\begin{itemize}
\item GeDi trained on sentiment of movie reviews can generate book text with a positive or negative tone better than state of the art baselines [Section \ref{sec:sent}]. Guiding towards positivity also has potential applications towards making LMs friendlier. 

\item GeDi is able to significantly reduce the toxicity of GPT-2 generation [Section \ref{sec:detox}], without sacrificing linguistic quality as compared with generating from GPT-2 directly, suggesting applications towards safer language modeling.

\item GeDi trained on a dataset of only 4 topics can generalize to new control codes zero-shot [Section \ref{sec:topic}], allowing them to guide generation towards a wide variety of topics. 


\item GeDi is very computationally efficient for both training and inference. GeDi guided generation in our experiments is more than $30\times$ faster than applying PPLM with GPT2-XL using default settings from \citet{dathathri2020plug}. Additionally, smaller GeDis fine-tuned for less than a day on a single GPU are effective and computationally efficient for controlling larger language models. This provides a cheap alternative to finetuning large LMs directly \citep{ziegler2019fine}.

\end{itemize}

\section{Background}
\subsection{Language modeling}
\label{clm}
 Language models (LMs) rely on an auto-regressive factorization to perform density estimation and generation of language data. Auto-regressive sequence models with parameters $\theta$ assign a probability to a sequence $x_{1:T} = \{x_1,\dots,x_T\}$ by factorizing it using the chain rule as follows:
\begin{equation}
P_{\theta}(x_{1:T}) = \prod_{t=1}^{T} P_{\theta}(x_t|x_{<t}).
\end{equation}
Models can assign probabilities to sequences by iteratively predicting a distribution over the next token given the previous tokens. Generating from language models requires iteratively sampling from $P_{\theta}(x_t|x_{<t})$, and then feeding $x_t$ back into the model as input for the next step.

\subsection{Class-Conditional Language modeling}
Class-conditional language models (CC-LMs) such as CTRL \citep{keskar2019ctrl} are a way for language models to generate while conditioning on an \emph{attribute} variable. 
CC-LMs predict a probability distribution $P_{\theta}(x_{1:T}|c)$, where $c$ is a class variable or a ``control code'' that describes an attribute of the text in $x_{1:T}$, which could, for instance, describe sentiment or topic. The auto-regressive factorization for a CC-LM is given by the following equation:
\begin{equation}
\label{eq:pctrl}
P_{\theta}(x_{1:T}|c) = \prod_{t=1}^{T} P_{\theta}(x_t|x_{<t},c).
\end{equation}
When training a CC-LM on a training set of sequences $\{x^{(1)}_{1:T_1},\dots,x^{(i)}_{1:T_i},\dots,x^{(N)}_{1:T_N}\}$, each sequence $x^{(i)}_{1:T}$ is paired with a control code $c^{(i)}$, which is a label or category of the sequence. The LM is trained to minimize the average negative log-likelihood, which we refer to as $\mathcal{L}_g$.
\begin{equation}
\label{eq:genloss}
\mathcal{L}_g  =  -\frac{1}{N}\sum_{i=1}^{N}\frac{1}{T_i}\sum_{t=1}^{T_i} \log P_{\theta}(x^{(i)}_t|x^{(i)}_{<t},c^{(i)}).
\end{equation}
In addition to class-conditional generation, CC-LMs can be used as generative classifiers by applying Bayes rule to compute $P_{\theta}(c|x_{1:T})$, as is done by \citet{keskar2019ctrl} for source attribution. 

\section{GeDi}
\label{sec:GeDi-LM}

\subsection{GeDi-guided Contrastive Generation}
\label{sec:contrastive}

\begin{figure}
\vspace{-1.3cm}
\begin{center}
\small

\includegraphics[width=0.7\textwidth]{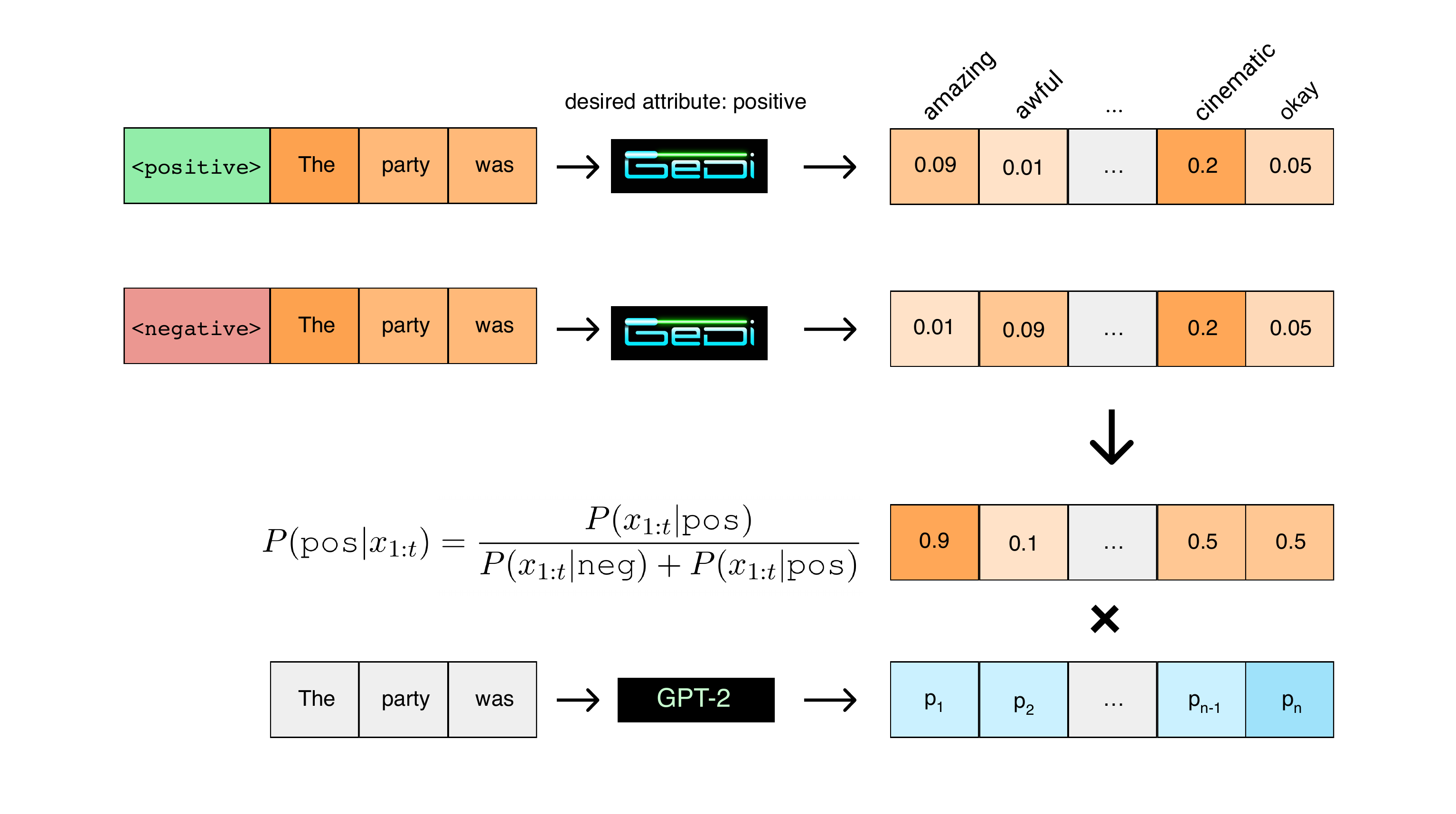}

\caption{\small A toy example of how GeDi-guided generation uses Bayes rule to efficiently compute classification probabilities for possible next tokens at each generation timestep using only element-wise operations.  These classification probabilities can then be used to guide generation from a language model (e.g., GPT-2) to achieve attribute control across domains. If the GeDi was trained on movie reviews for sentiment control, its direct class-conditional predictions will be biased towards predicting movie review words (illustrated by next word prediction of ``cinematic''). However, by contrasting the predictions of opposing control codes via Bayes rule, the bias towards movie reviews can be canceled out.}
\vspace{-1em}
\label{fig:figure_1}
\end{center}
\end{figure}

GeDi assumes we have a CC-LM with desired control code $c$ and an undesired or \emph{anti-control code} $\bar{c}$, and uses the contrast between $P_{\theta}(x_{1:t}|c)$ and  $P_{\theta}(x_{1:t}|\bar{c})$ to guide sampling from an LM that gives $P_{{LM}}(x_{1:t})$. Specifically, when predicting the next token during generation, GeDi uses this contrast to compute the probability that every candidate next token $x_t$ belongs to the desired class, given by $P_{\theta}(c|x_t,x_{<t})$. Our key insight is that this distribution can be computed very efficiently when using CC-LMs as GeDis via application of Bayes rule for partial sequences during generation.

\begin{equation}
\label{eq:ponline}
P_{\theta}(c|x_{1:t}) = \frac{P(c)\prod^t_{j=1} P_{\theta}(x_{j}|x_{<j},c)}{\sum_{c' \in \{c, \bar{c}\}} \prod^t_{j=1} P(c')P_{\theta}(x_{j}|x_{<j},c')}.
\end{equation}

When computing this online during sequence generation, the model will have already computed $P_{\theta}(x_{j}|x_{<j},c')$ for any $j<t$ from the previous time-steps, and it will only need to compute $P_{\theta}(x_{t}|x_{<t},c')$. This can be computed in two parallel forward passes; one conditioning on $c$ and one conditioning on $\bar{c}$ (both conditioning on the same $x_{<t}$). The model can also save the hidden states from the previous time steps to avoid computing a forward pass through the full sequence at each next token generation step. Applying a unidirectional classifier such as GPT \citep{radford2018improving} to compute $P_{\theta}(c|x_t,x_{<t})$ directly  (i.e. discriminatively) would require feeding in every possible input $x_t \in \gV$ into the classifier, and thus would require $|\gV|$ forward passes for a vocab set $\gV$. A bidirectional classifier such as BERT \citep{devlin2018BERT} would require $t \times |\gV|$ forward passes because it would need to recompute attention states from earlier time-steps. For typical vocab sizes of $20$k+, GeDi's online classification trick can compute $P_{\theta}(c|x_t,x_{<t})$ for every possible next token $x_t$ on the order of 10k fold less computation as compared with a unidirectional classifier. 

In practice, we find normalizing ($\log$) probabilities by current sequence length $t$ results in more robust generation of variable length sequences. Our GeDi trained models (see next section) also use a learnable scale parameter $\alpha$. To compute $P_{\theta}(c|x_{1:t})$ for GeDi-guided generation, we use the following equation:

\begin{equation}
\label{eq:GeDi}
P_{\theta}(c|x_{1:t}) =  \frac{P(c) \, P_{\theta}(x_{1:t}|c)^{\alpha/t}}{\sum_{c' \in \{c, \bar{c}\}}  P(c') \, P_{\theta}(x_{1:t}|c')^{\alpha/t}}.
\end{equation}

The log prior is encoded with bias parameters $b_c$, where $P(c) = \frac{e^{b_c}}{\sum_{c'} e^{b_{c'}}}$. This bias parameter can be assumed to be zero for uniform classes, learned (see next section on GeDi training), or set manually as a hyper-parameter. In practice, $P_{\theta}(c|x_{1:t})$ is computed with log-probabilities (see Appendix \ref{apdx:logGeDi}).  With the efficient estimation of $P_{\theta}(c|x_t,x_{<t})$, there are many possible heuristics that can be used to guide LM generation, so long as the LM and GeDi share the same tokenization. Heuristics that use $P_{\theta}(c|x_t,x_{<t})$ inherently contrast predictions conditioned on $c$ and $\bar{c}$, causing attributes common to $c$ and $\bar{c}$ to be cancelled out, more effectively allowing for the attribute described by $c$ to be transferred across domains, as illustrated in Figure \ref{fig:figure_1}. 

\subsubsection{Heuristics for guiding generation}
We applied weighted decoding and filtering heuristics to use $P_{\theta}(c|x_t,x_{<t})$ to guide generation, which worked well in practice in our experiments but are not necessarily optimal; there are many possible ways to use the classification signal given by GeDi to guide generation. Our initial heuristic applies a weighted posterior given by
\begin{equation}
\label{eq:posterior}
P_w(x_t|x_{<t},c) \propto \ P_{{LM}}(x_t|x_{<t})P_{\theta}(c|x_t,x_{<t})^\omega ,
\end{equation}
where $\omega>1$ to bias generation more strongly towards the correct class. The right hand side of Equation (\ref{eq:posterior}) is normalized over all $x_t$ in the vocabulary to obtain $P_w(x_t|x_{<t},c)$.  

While we found that the weighted posterior in Equation (\ref{eq:posterior}) is most critical for controlling generation, we also used an additional filtering heuristic that was beneficial for steering generation more aggressively. This heuristic, inspired by \emph{nucleus sampling} \citep{holtzman2019curious}, removes candidate next word tokens with lower values for $P_{\theta}(c|x_t,x_{<t})$ while maintaining a minimum of at least $\rho$ in cumulative probability mass in  $P_w(x_t|x_{<t},c)$. We define $\gV_n$ as the set of $n$ tokens with the highest $P_{\theta}(c|x_t,x_{<t})$. We define $m$ as the minimum $n$ such that
\begin{equation}
\sum_{x_t \in \gV_n} P_w(x_t|x_{<t},c) \geq \rho . 
\end{equation}
We define $\gV_m$ as $\gV_n$ for $n=m$, meaning that $\gV_m$ will contain the minimum number of tokens possible at the head of the distribution for $P_{\theta}(c|x_t,x_{<t})$ to maintain a minimum cumulative probability of $\rho$ in $P_w(x_t|x_{<t},c)$.

We define another set of tokens to keep, $\gV_p \subseteq \gV$, which maintains all tokens where $P_{\theta}(c|x_t,x_{<t})>\tau$. The motivation is that if we are acceptably sure that the resulting sequence from generating a token is in the correct class, there is no need to filter it. The final set of tokens to keep are then given by $\gV_k = \gV_p \cup \gV_m$. We then zero out probabilities of tokens not in $\gV_k$ and re-scale the remaining distribution to sum to $1$.

\subsection{GeDi Training}
 \label{sec:training}
The previous section presented a method for using a CC-LM as a GeDi to guide the generation of another LM. However, previous work shows that generative classifiers are generally inferior to discriminative ones when trained on large datasets \citep{ng2002discriminative,yogatama2017generative}. For this reason, we propose training CC-LMs discriminatively as classifiers with GeDi training, with the primary goal of making them better discriminators for GeDi-guided generation. We also have a secondary goal of making them better at directly generating; a CC-LM that can correctly classify sequences via Equation (\ref{eq:GeDi}) may be better at generating sequences in the desired class. The idea of discriminatively training class-conditional generative models has previously been considered for the classification of text \citep{yakhnenko2005discriminatively}, and images \citep{lasserre2006principled}.

With GeDi training, we combine the standard generative language modeling loss  $\mathcal{L}_g$ from Equation (\ref{eq:genloss}) with a discriminative loss $\mathcal{L}_d$,  defined as: 
\begin{equation}
\mathcal{L}_d  = -\frac{1}{N} \sum_{i=1}^{N} \log P_{\theta}(c^{(i)}|x^{(i)}_{1:T_i}). 
\end{equation}
$P_{\theta}(c^{(i)}|x^{(i)}_{1:T_i})$ is derived from an offline version of Equation (\ref{eq:GeDi}) given by

\begin{equation}
\label{eq:GeDi_train}
P_{\theta}(c^{(i)}|x^{(i)}_{1:T_i}) =  \frac{P(c) \, P_{\theta}(x_{1:T_i}^{(i)}|c^{(i)})^{\alpha/T_i}}{\sum_{c'}  P(c') \, P_{\theta}(x_{1:T_i}^{(i)}|c'^{(i)})^{\alpha/T_i}} ,
\end{equation}

where $c' \in\{c^{(i)},\bar{c}^{(i)}\}$ for the binary case (where $c^{(i)}$ is the correct class and $\bar{c}^{(i)}$ is the incorrect class for the $i$th sequence), $P(c) = \frac{e^{b_c}}{\sum_c' e^{b_{c'}}}$ (where $b_c$ is a learnable class bias which we omit when class distribution is roughly equal), $\alpha$ is a learnable scale parameter, and $P_{\theta}(x_{1:T_i}^{(i)}|c^{(i)})$ is given by Equation (\ref{eq:pctrl}) for CC-LMs. The cost function for GeDi training $\mathcal{L}_{gd}$ is then given by 
\begin{equation}
\label{eq:hybrid_obj}
\mathcal{L}_{gd} = \lambda \mathcal{L}_g + (1-\lambda) \mathcal{L}_d,
\end{equation}
where $\lambda$ is a hyper-parameter. In GeDi training, the discriminative loss $\mathcal{L}_d$ is aimed at increasing classification accuracy, whereas the generative loss $\mathcal{L}_g$ likely helps the CC-LM have better calibrated token probabilities for guided generation.

\section{Related Work}
\label{sec:related}

Methods for controlling text generation can be categorized broadly into two types:
training or finetuning a model directly for controllable generation~\citep{keskar2019ctrl,ziegler2019fine,rajani2019explain,ficler2017controlling,yu2017seqgan,Hu2017TowardCG} or using a discriminator to guide generation ~\citep{ghazvininejad2017hafez,holtzman2018learning,dathathri2020plug}.
\citet{keskar2019ctrl} train a CC-LM with pre-defined control codes placed at the start of every sequence. Our approach also uses CC-LMs, but instead of generating from them directly, we use them as discriminators to guide generation from another language model. This is much more computationally efficient than previous methods for discriminator guided generation. \citet{holtzman2018learning} apply discriminators to re-weight a beam search, requiring all candidate tokens to be passed through the discriminator, scaling linearly with the number of re-scored tokens. PPLM \citep{dathathri2020plug} trains an attribute model on top of a language model's last hidden layer and backpropagates gradients to update the hidden states of the model. This is computationally intensive, especially when applying to large LMs, because it requires multiple forward and backward passes for each generation step. 

GeDi also relates to contrastive learning \citep{smith-eisner-2005-contrastive,Mnih-nce-icml-12}. Most existing contrastive learning methods work at the instance level by constrasting one positive pair from $k$ negative pairs, whereas GeDi works at the class level and contrasts a positive class-conditional distribution against a negative one. GeDi also uses the contrast between positive and negative distributions for both training (i.e., GeDi training) and inference (i.e., contrastive generation).

\section{Experiments}
\label{sec:discriminative}

Our experiments finetune GPT2-medium (345M parameter) \citep{radford2019language} with control codes specific to each task to form a class-conditional language model. We consider finetuning using GeDi training ($\lambda < 1$ in Equation (\ref{eq:hybrid_obj})) and standard generative training ($\lambda = 1$ in Equation (\ref{eq:hybrid_obj})). These experiments were performed using adaptations of Huggingface Transformers \citep{wolf2019transformers}. We study the trade-offs between GeDi vs generative training for classification, perplexity, and direct generation in depth in Appendix \ref{sec:generative}. We find that GeDi trained CC-LMs have a higher generative classification accuracy at the cost of a higher perplexity. We also find that GeDi-trained CC-LMs are able to achieve a higher label fidelity across generation tasks, meaning that the control code more often corresponds to the true attribute of the generated sample. 

In our main experiments, we use these CC-LMs as GeDis to guide generation from GPT2-XL (1.5B parameter). For generation, we use greedy decoding with a repetition penalty \citep{keskar2019ctrl}, and condition on varying prompts to give diversity across samples. Additional details about the way we apply a repetition penalty are given in Appendix \ref{apdx:gen_settings}, and our hyper-parameter settings for GeDi-guided generation, which were shared across most experiments, are given in Appendix \ref{apdx:gedi-hyper}. We experiment with GeDi-guided generation for sentiment, detoxification, and topic control.   
 
 In our sentiment experiments, we compare direct generation from CC-LMs vs. using CC-LMs as GeDis. We refer to direct generation simply as ``CC-LM'' (using $\lambda=1$ to specify generative training and $\lambda<1$ to specify GeDi training), and we refer to GeDi-guided generation using a CC-LM to guide GPT-2 as ``GeDi-guided'' (also using $\lambda$ to specify generative/GeDi training).
 
\subsection{Guiding sentiment control across domains}
\label{sec:sent}
We experiment with GeDi-guided generation from GPT-2 for sentiment control. For these experiments, we use CC-LMs finetuned on IMDb movie reviews using both GeDi and generative training (reused from Appendix \ref{sec:generative}). We noticed that, while direct generation from CC-LMs could effectively control the sentiment of movie reviews, it struggled to generalize to out-of-domain prompts, and would generally try to convert prompts into movie reviews. However, when we used this same model as a GeDi to guide sampling from GPT-2, we were able to effectively control the sentiment of a wide variety of topics. For instance, in our preliminary experiments, we considered the prompt ``I just read this paper on Generative-Discriminative training.'' in Table \ref{gedi_disc_sentiment_deep_learning} and it results in text that mentions well known deep learning ideas and researchers while also controlling sentiment.

\begin{wraptable}[11]{r}{5.5cm}
\scriptsize
\caption{\small Average generation time in seconds per token for generating sequences of length 256. }
\begin{center} 
\begin{tabular}{ @{} l  l@{} } \toprule 
Model & Generation time  \\
& (sec/token)\\
\midrule 
GPT2-XL & 0.060   \\
GeDi-guided (w/\, GPT2-XL) & 0.095   \\
PPLM (w/\, GPT2-XL) & 3.116 \\
\bottomrule
\end{tabular} 
\end{center}

\label{tab:gen_times}
\end{wraptable}

\normalsize
To experimentally verify that GeDi can achieve domain transfer of the concepts of ``positivity'' and ``negativity'', we consider a book text generation task where we conditionally generate text from the start of book chapters from Bookcorpus \citep{Zhu_2015_ICCV}, where each prompt is at least 150 characters and ends on the first-word break after the minimum length. We run human evaluation on generations from 50 different book prompts from 13 different models; including raw GPT2-XL, and the following models with both positive and negative sentiment: 
\begin{enumerate*}
    \item GPT2-XL guided by a GeDi-trained CC-LM (GeDi-guided, $\lambda=0.6$),
    \item GPT2-XL guided by a generatively-trained CC-LM (GeDi-guided, $\lambda=1.0$),
    \item direct generation from a GeDi-trained CC-LM (CC-LM, $\lambda=0.6$),
    \item direct generation from a generatively-trained CC-LM (CC-LM, $\lambda=1.0$),
    \item CTRL,
    \item PPLM applied to GPT2-XL.
\end{enumerate*}
See Appendices \ref{apdx:pplm-hyper} and \ref{apdx:ctrl-hyper} for additional information about our PPLM and CTRL baselines respectively. We found that it was more than $30\times$ faster to guide GPT2-XL with a GeDi as compared with PPLM (assuming 10 update steps as used in \cite{dathathri2020plug}), as shown in Table \ref{tab:gen_times}).

Amazon Mechanical Turk annotators rated the generated text on sentiment/tone, how book-like the text was, and whether or not the text resembled an Amazon review or movie review (since CTRL was trained on Amazon reviews and GeDi was trained on movie reviews). Each annotator was randomly assigned samples from the set of all generations from all models. The results are given in Table \ref{tab:booksent}. Using a GeDi-trained CC-LM to guide GPT2-XL was able to generate book-like text while strongly control the tone. GeDi was also able to give slightly stronger sentiment control than PPLM, in addition to being more than $30\times$ faster.

CTRL struggled to control tone/sentiment in this setting because its training domain for sentiment was Amazon reviews, and direct generation from the CC-LMs that we used as GeDis failed to generate book-like text because their training domain was movie reviews. We provide examples of generations from all models on book prompts in Appendix \ref{apdx:sent}. Table \ref{tab:ctrl_gedi_gen_book_prompt_sentiment} specifically shows how CTRL tends to generate Amazon reviews and how the generative and GeDi-trained CC-LMs tend to generate movie reviews. Using these same CC-LMs as GeDis to guide generation led to book-like text, demonstrating domain transfer of the concepts of positivity and negativity. 

\begin{table*}[t]
\caption{\small Human evaluation for sentiment on book text generation (rated for positivity and book resemblance both on a scale of 1-5), with key results in \textbf{bold}. We average two annotations on generations from 50 prompts for each model, where prompts are from the start of book chapters, and are a minimum of 150 char. We compare using a CC-LM as a GeDi to guide GPT2-XL (GeDi-guided), vs. direct class conditional generation (CC-LM). We also compare GeDi trained CC-LMs ($\lambda=0.6$) vs. generatively trained CC-LMs ($\lambda=1.0$) for both types of generation methods, with both positive (pos) and negative (neg) control codes. The GeDi-trained GeDi guide (GeDi-guided-neg ($\lambda=0.6$) and GeDi-guided-pos ($\lambda=0.6$)) was able to reliably control sentiment while also generating book-like text, even though the CC-LM used as a GeDi was trained on movie reviews. Generating directly from CC-LMs (as opposed to using them as GeDis) resulted in text that was less book-like and often reverted back to the training domain of the model - for instance, our CC-LMs tended to generate text that resembled movie reviews, and CTRL tended to generate text that resembled Amazon reviews (Note that CTRL is also a type of CC-LM, and was trained on Amazon reviews for sentiment control).} 
\begin{center} 
\small
\begin{tabular}{  l  c c c c} \toprule 
Model & Positivity & Book-like $\uparrow$ & Movie review $\downarrow$ & Amazon review $\downarrow$ \\
\midrule 
GeDi-guided-pos ($\lambda=1.0$) & \textbf{3.85} & \textbf{4.11} & 2 \% & 9 \% \\
GeDi-guided-pos ($\lambda=0.6$) & \textbf{3.65} &  \textbf{4.19} & 2 \% & 1 \% \\ 
PPLM-pos & 3.53 & 4.14 & 3 \% & 3 \% \\
CC-LM-pos ($\lambda=1.0$) & 3.13 & 2.86 & \textbf{55 \%} & 9 \% \\
CC-LM-pos ($\lambda=0.6$) & 3.36 & 2.92 & \textbf{61 \%} & 11 \% \\
CTRL-pos & 2.86 & 3.81 & 10 \% & \textbf{29 \%} \\
\midrule 
GPT2-XL & 3.18  & 4.18 & 3\% & 8\% \\
\midrule
CTRL-neg & 2.90 & 3.64 & 4\% & \textbf{32\%} \\
CC-LM-neg ($\lambda=0.6$) & 2.15 & 2.68 & \textbf{65\%} & 8 \% \\
CC-LM-neg ($\lambda=1.0$) & 2.50 & 2.84 & \textbf{63\%} & 8 \% \\
PPLM-neg & \textbf{2.62} & 3.96 & 2\% & 5\% \\ 
GeDi-guided-neg ($\lambda=0.6$) & \textbf{1.98} & \textbf{4.02} & 7\% & 8 \% \\
GeDi-guided-neg ($\lambda=1.0$) & \textbf{1.85} & \textbf{3.62} & 16\% & 7 \% \\
\bottomrule
\end{tabular} 

\end{center}

\label{tab:booksent}
\end{table*}

\subsection{Detoxifying GPT-2}
\label{sec:detox}
With the motivation of detoxifying GPT-2, we train a CC-LM as a toxicity classifier on the Jigsaw Toxic Comment Classification Challenge Dataset\footnote{\scriptsize\url{https://www.kaggle.com/c/jigsaw-toxic-comment-classification-challenge/}}, which contains text samples labeled as ``toxic'' or ``non-toxic''. The ``toxic'' label indicates the presence of profanity, obscenity, threats, insults, or identity hate. We train models on an even split of toxic and non-toxic examples. We use toxic examples from the Jigsaw dev set to find prompts to condition on for evaluation. We used prompts that ended on a word break and were at least 30 characters.  In order to have prompts that are more likely to trigger aggressive generations but less likely to be explicitly toxic, we pass candidate prompts through a RoBERTa \citep{liu2019roberta} model trained to classify toxicity, and only kept prompts where RoBERTa was less confident about the toxicity label. We generate samples from these prompts using GeDi-guided generation with a GeDi-trained guide ($\lambda=0.6$) and a generatively trained guide ($\lambda=1.0$). 

\begin{wraptable}[11]{r}{6.6cm}
\caption{\small Average toxicity (scale of $1$-$3$) and linguistic quality scores  (scale of $1$-$4$) for $100$ samples for each model. Both the GeDi-trained GeDi guide ($\lambda=0.6$) and generatively-trained GeDi guide ($\lambda=1.0$)  resulted in significantly less toxic text as compared with GPT2-XL without sacrificing linguistic quality.}
\scriptsize
\centering
\begin{tabular}{@{}l c c@{}} \toprule 
Model & Toxicity $\downarrow$  & Linguistic quality $\uparrow$ \\
\midrule 
GPT2-XL & 1.45  & 3.23  \\
\midrule
GeDi-guided ($\lambda=0.6$)  & 1.17  & 3.44  \\
GeDi-guided ($\lambda=1.0$)  & 1.13  & 3.25  \\
\bottomrule
\end{tabular} 
\label{tab:detox_humanfid}
\end{wraptable}

We run human evaluation to measure toxicity  [1: non-toxic, 2: mildly toxic, 3: toxic] and linguistic quality  [1: very low quality, 2: low quality, 3: high quality, 4: very high quality].  Results are given in Table \ref{tab:detox_humanfid}. GeDi-guided generation resulted in significantly less toxic text for both values of $\lambda$, with the  GeDi-trained GeDi guide ($\lambda=0.6$) achieving the highest linguistic quality of all models.

\subsection{Multi-class topic control}
\label{sec:topic}
\begin{wraptable}[16]{r}{7.5cm}
\centering
\scriptsize
\caption{\small Average topic relevance (reported on a scale of $1$-$5$, where higher is more relevant) and linguistic quality scores (scale of $1$-$4$)  for $100$ samples from each model for each of the four topics {Business, Science/Tech, Sports, World}. GeDi was able to control topic while maintaining a similar level of linguistic quality to GPT-2. The GeDi guide was trained on AG-news using GeDi training ($\lambda=0.8$). }
\begin{tabular}{@{}c c c c@{}}
\toprule 
Topic & Model & Relevance $\uparrow$  & Linguistic quality $\uparrow$ \\

\midrule
Business & GPT2-XL & 1.95 & 3.44 \\
 & GeDi-guided  & 4.41  & 3.21  \\
\midrule
Science/Tech & GPT2-XL & 1.97 & 3.63 \\
 & GeDi-guided  & 3.45  & 3.58  \\
\midrule
Sports & GPT2-XL & 1.31 & 3.49 \\
 & GeDi-guided  & 3.81  & 3.44  \\
\midrule
World & GPT2-XL & 2.75 & 3.44 \\
 & GeDi-guided    & 3.99  & 3.39  \\
\bottomrule
\end{tabular} 
\label{tab:topic_humanfid}
\end{wraptable}

We extend GeDi to the multi-class setting by training it to classify whether or not a sequence matches a topic. This can be done with CC-LM by training it to condition on a ``true'' and ``false" control code, where sequences have the name of a topic prepended. The ``true'' control code corresponds to sequences where the prepended topic matches the sequence, and the ``false'' control code corresponds to sequences where the prepended topic does not match the text. For generation, the desired attribute is set to ``true'' and the prompt is prepended with the desired topic. Refer to Appendix \ref{sec:multi} for additional details. We use the AG news topic classification data set \citep{zhang2015character} which has 4 topics (World, Sports, Business, and Science/Tech) to train GeDis with 6 values of $\lambda$ (including $\lambda=1$). We only train the CC-LMs on half of the dataset and train a RoBERTa classifier on the other half to measure label fidelity.  After training, we applied each GeDi to guide generation from GPT2-XL. We use a minimum of 50 character prompts from the multi-news dataset \citep{fabbri2019multi} to condition on for generation. The prompt often will not fit with the desired topic, sometimes creating a challenge for the model to relate the article to the desired attribute. We measured automatic label fidelity first as given by the RoBERTa classifier, and we found the generatively trained GeDi guide  ($\lambda=1$) achieved a significantly lower automatic label fidelity (61\% for $\lambda=1$ vs. 74\% for $\lambda=0.8$), suggesting that GeDi-training may be important for extending GeDi-guided generation to many control codes using the proposed binarization method.

We ran human evaluation on samples from the 4 news topics comparing our strongest GeDi guide (we chose $\lambda=0.8$ based on automatic label fidelity), and raw GPT-2-XL. Annotators were given the topic and asked to rate samples on topic relevance and linguistic quality. The results are given in Table \ref{tab:topic_humanfid}. GeDi-guided generation gave text with high relevance for all 4 topics while maintaining a similar level of linguistic quality to GPT2-XL. We give examples of GeDi topic generation in Appendix \ref{apdx:topic}.

\subsubsection*{Zero-shot control codes}
\label{sec:zeroshot}
For topic training, we prepended the words ``world'', ``sports'', ``business'', and ``science'' to sequences. However, at inference, any word could potentially be prepended to the prompts. We observed that the GeDi, trained for only several hours on a single GPU on 4 topics, could guide GPT-2 towards generating text corresponding to a very wide array of topics that included ``space'', ``history'', ``education'', ``cars'', ``climate'' and many more. This zero-shot behavior worked very well for short, topic neutral prompts, as shown for the prompt ``In a shocking finding'' in Appendix \ref{apdx:zeroshot}, but did not work as well for longer prompts.  We also only tested topics that could be encoded with 1 byte-pair encoding \citep{sennrich2015neural} token, since this was the case for all our training topics. However, this zero-shot behavior could likely apply to longer control codes if trained on longer control codes. We also compare with zero-shot topic generation using CTRL in Table \ref{ctrl_shocking_finding} as a baseline and find that despite being trained on significantly more topics, CTRL struggles to generate text corresponding to control codes it has never seen during training. 

GeDi's ability to generalize to new control codes zero-shot gives the ability to generate text corresponding to many topics and subtopics. This ability likely emerges because generative classifiers can classify unseen topics zero-shot from learned word embeddings \citep{yogatama2017generative}, and GeDi uses generative classifiers to guide generation. This is another advantage of GeDi over the previous discriminator guided generation approaches.

\section{Future directions}
Methods to make large LMs like GPT-3 \citep{brown2020language} safer and more controllable are becoming especially important as LMs become incorporated into products. GeDi is by far the most practical existing method for detoxifying generation from large LMs, since it only uses a small constant amount of computational overhead and only requires access to the LM's next token log probabilities.  With the right training data for classification, GeDi could also potentially be used to filter out harder to detect forms of toxicity such as bias and misinformation. Extending on the methods in this paper, multiple GeDis trained to filter out different undesirable attributes could be combined, for instance by multiplying the attribute classification terms from several different discriminators in Equation \ref{eq:posterior}. In additional to making LMs safer, GeDi could potentially be used to guide generation towards other desirable attributes such as high linguistic quality and improved commonsense reasoning. Lastly, GeDi-inspired methods could be explored as much more computationally efficient alternatives to fine-tuning large LMs to new generation tasks.

\section{Conclusion}

We present GeDi as an approach for controllable generation that uses generative discriminators to classify candidate next tokens on the fly during inference, making it far more efficient than previous methods that use discriminators to guide generation. GeDi achieves stronger controllability of sentiment than PPLM while also giving a generation speed more than $30\times$ faster.  GeDis trained on 4 topics can also controllably generate new topics zero-shot from just a key word, unlocking a new capability that previous controllable generation methods like PPLM and CTRL do not have. We also show that GeDi is able to significantly reduce the toxicity of GPT-2 without sacrificing linguistic quality. The ethical considerations of language modeling are becoming more important as LMs like GPT-3 become incorporated into products, and GeDi is far more promising than any previous approach for detoxifying large language models while maintaining a fast generation speed. This work also moves towards unifying natural language generation with classification, and suggests that we may be able to efficiently generate text that corresponds to any attribute that we can accurately classify. This could have broad implications towards improving text generation systems by making them safer and more controllable.

\subsubsection*{Author Contributions}
Ben thought of the main ideas and designed the research. Ben and Akhilesh coded the implementation. Akhilesh maintained the codebase, set up automatic and human evaluation experiments, and organized results. Nazneen advised on detoxification experiments. All authors contributed to writing and discussions.

\subsubsection*{Acknowledgments}
The authors thank Semih Yavuz and Yu Bai for helpful discussions and feedback on this project.

\bibliography{iclr2021_conference}
\bibliographystyle{iclr2021_conference}

\clearpage

\appendix

\subfile{appendix}

\end{document}

%% file: math_commands.tex
\usepackage{amsmath,amsfonts,bm}

\def\eqref#1{equation~\ref{#1}}

\def\1{\bm{1}}

\DeclareMathAlphabet{\mathsfit}{\encodingdefault}{\sfdefault}{m}{sl}
\SetMathAlphabet{\mathsfit}{bold}{\encodingdefault}{\sfdefault}{bx}{n}

\def\gV{{\mathcal{V}}}



%% file: appendix.tex
\section{Multi-class GeDi}
\label{sec:multi}

Both GeDi-guided generation and GeDi training use CC-LMs to perform classification. The most straightforward way to extend this to many classes is to have one forward pass conditioned on each control code and normalize over a larger number of classes via Equation (\ref{eq:GeDi}) (which we in fact do for 3-class MNLI in Appendix \ref{sec:generative}). However, this approach does not scale well computationally to large numbers of classes. As a solution, we propose reframing each classification task as binary classification using control codes and anti control codes for each class. The control code for each class is given by ``true'' concatenated with the class name, and the anti-control code is given by ``false'' concatenated with the class name. The CC-LM then classifies whether the class name corresponds to the text. For instance, the CC-LM would process the following two sequences in parallel: 

\noindent
\scriptsize
\texttt{\textbf{\textless true\textgreater \, \textless science\textgreater} \, T-rex achieved its massive size due to an enormous growth spurt during its adolescent years.}
\\
\\
\texttt{\textbf{\textless false\textgreater \, \textless science\textgreater} \, T-rex achieved its massive size due to an enormous growth spurt during its adolescent years.}
\normalsize

\noindent and would classify it as \emph{true} or \emph{false} as to whether the class (in this case ``science'') matches the category of the text by using Equation (\ref{eq:GeDi_train}). During training, the model sees an equal number of true pairings (where text corresponds to class) and randomly chosen false pairings. After the model has been trained, binary GeDi-guided generation can be applied, using $c=$\small \texttt{\textless true\textgreater} \normalsize and $\bar{c}=$\small \texttt{\textless false\textgreater},\normalsize \, and using the desired class name as the first token ($x_1$) in the sequence. This also makes it possible to form new control codes zero-shot; a new topic word that was never seen before in training can be chosen in place of $x_1$.

\normalsize

\section{GeDi with log probabilities}
\label{apdx:logGeDi}

GeDi-guided generation and GeDi training both use language models discriminatively via Bayes rule by using

\begin{equation}
P_{\theta}(c|x_{1:T}) =  \frac{P(c) \, P_{\theta}(x_{1:T}|c)^{\alpha/T}}{\sum_{c'}    P(c') \, P_{\theta}(x_{1:T}|c')^{\alpha/T}} ,
\end{equation}
where
\begin{equation}
P(c) = \frac{e^{b_c}}{\sum_{c'} e^{b_{c'}}}.
\end{equation}

For GeDi-guided generation, this is computed online for partial sequences during generation, whereas for GeDi training, it is computed for full training sequences. For numerical stability, we compute this using log-probabilities. Log-probabilities for each class are given by


\begin{equation}
\label{eq:GeDi_softmax2}
\log P_{\theta}(x_{1:T}|c)  =  \sum_{t=1}^{T}  \log P_{\theta}(x_t|x_{<t},c) ,
\end{equation}
and the class probability is given by 
\begin{equation}
\label{eq:GeDi_softmax3}
P_{\theta}(c|x_{1:T}) =  \frac{e^{({b_c + (\alpha/T) \log P_{\theta}(x_{1:T}|c)})}}{\sum_{c'} e^{(b_{c'} + (\alpha/T) \log P_{\theta}(x_{1:T}|c'))}} .
\end{equation}
This can be computed in a numerically stable way using a softmax \citep{bridle1990probabilistic}, since the maximum logit to the softmax can be subtracted out before taking the exponent without changing the result. For the two class case (all of our experiments except for MNLI, which was 3-class), $c'\in \{c,\bar{c} \}$, meaning that the above equation could have been equivalently computed using a sigmoid of the difference of the logs of the two terms in the denominator sum (but our implementation used softmax as above). 

\section{Generation settings}
\label{apdx:gen_settings}
When comparing the quality of samples from different language models, there is a trade-off between quality and diversity; models that tend to have more sharply peaked distributions for $P_{\theta}(x_t|x_{<t},c)$ will tend to have higher quality samples, but will also have less diversity. Applying GeDi results in more sharply peaked distributions due to the filtering step, which zeros out probabilities for some tokens. In order to ensure a fair comparison of models, we only use greedy decoding for our experiments, meaning we always pick the most likely token in the model's predictive distribution. With greedy decoding, the model would generate the same text sequence every time without any conditioning text. Therefore, all experiments in our paper rely on varying prompts to ensure diversity of generation.

We also apply a repetition penalty~\citep{keskar2019ctrl}, which we found necessary for preventing degeneration with greedy decoding. Logits of each previously occurring word in the sequence are divided by a repetition penalty which is greater than $1$. To account for the possibility of negative logits, we re-scaled the final logits in all models to always have a maximum of $10$ across the vocabulary before dividing by the repetition penalty. We used a repetition penalty of $1.2$ in all models in our experiments.

\section{Additional model and hyper-parameter details}
\label{apdx:sentbaselines}

\subsection{Hyper-parameters for GeDi guided generation}
\label{apdx:gedi-hyper}
We found hyper-parameter settings using a combination of eyeballing generation quality and automatic label fidelity metrics given by a RoBERTa classifier \citep{liu2019roberta} trained on an external training set (where a label fidelity of 100\% means that RoBERTa always agrees that the class label is the same as the control code). All of our GeDi models except for the GeDi-trained detoxification model use the same generation hyper-parameters ($\omega = 30$, $\rho=0.2$, $\tau=0.8$), which we found to work well across tasks and values of $\lambda$ for training. 

Using the hyper parameters above, we initially found that the GeDi-trained detoxification guide would sometimes result in very short samples that cut off mid sentence. Since the GeDi operates discriminatively at the token level, it cannot confidently classify a sequence as non-toxic until the sequence has finished, which likely was causing the model to finish sequences early to ensure that they would not become toxic. To fix this problem, we manually added a bias parameter $b_c=2$ as per Equation (\ref{eq:GeDi}) so that the model would have a prior probability that assumes the sequence is non-toxic. We found doing this also required us to increase $\tau$ to $0.97$ to account for $P(c|x_{1:t})$ being higher with the bias parameter, since otherwise far fewer tokens would be filtered and samples could become toxic. All other hyper-parameters remained unchanged.

\subsection{Baseline details for PPLM}
\label{apdx:pplm-hyper}
For PPLM, we trained the external classifier (which uses logistic regression on top of representations from GPT-2) on the SST-5 data set, after struggling to achieve as strong results training on IMDb (which is what GeDi was trained on) and advise from the paper authors. For generation, we used greedy decoding with a repetition penalty applied the same way as described in Appendix \ref{apdx:gen_settings}. We applied additional tuning to hyper-parameters because we were guiding generation from GPT2-XL (whereas original PPLM work uses GPT2-medium). Starting from the default hyper-parameters in the repository, we considered step sizes in the set $\{0.04, 0.08, 0.16, 0.25, 0.35\}$, and found that $0.25$ gave the best trade-off between sentiment control and generation quality, so we used this for our experiments.

\subsection{Baseline details for CTRL}
\label{apdx:ctrl-hyper}
For CTRL, we prepended prompts with the control codes for positive and negative Amazon reviews, which are ``Reviews Rating: 1.0'' and ``Reviews Rating: 5.0'' for negative and positive respectively. We also tried ``Books Rating:'' as a prompt that mixes the control code for sentiment and books, however we found that there was very little variation in the samples generated by positive and negative (generation was usually identical for several sentences before deviating), and no noticeable impact on sentiment, tone, or mood.

\clearpage

\section{Experiments with GeDi training}
\label{sec:generative}
Our initial experiments train and benchmark GeDi-trained CC-LMs for classification, perplexity, and direct generation, in preparation to use them for GeDi-guided generation in Section \ref{sec:discriminative}. All our experiments augment GPT2-medium (345M parameter) \citep{radford2019language} with control codes specific to each task to form a class-conditional language model.
We then fine-tune this model on different sequence classification datasets with the hybrid GeDi objective from Equation (\ref{eq:hybrid_obj}).
To understand the trade-offs between generative and discriminative training, we explore $\lambda$ values between $0$ and $1$, 
where $\lambda=1$ is equivalent to generative training and is the main baseline for these initial experiments. 
Once fine-tuned, 
we decode samples from the model by conditioning on the control code corresponding to the required attribute and prompts from the dev set for each task. We use greedy decoding and a repetition penalty for generation (see Appendix \ref{apdx:gen_settings} for details) On each task, we measure the perplexity, classifier accuracy, and label fidelity across all values of $\lambda$. Our task set consists of:

\textbf{IMDb} \citep{maas-EtAl:2011:ACL-HLT2011}: We test the model's ability to generate movie reviews with positive and negative sentiment when conditioned on the first $\sim$100 characters (up to the next word-break after 100 characters) of a review (which may or may not match the control code).

\textbf{MNLI} \citep{williams2017broad}: We test the model's ability to generate contradictions and entailments when conditioned on a premise.

\textbf{QNLI} \citep{wang2018glue}: We test the model's ability to generate passages that contain the answers to a question given in conditioning text.

We include the two NLI tasks because they require a greater degree of logical reasoning, potentially making them more difficult.

\subsection{Evaluation of GeDi-trained CC-LMs}
\label{subsec:modeleval}
To evaluate the label fidelity of direct generation from GeDi-trained CC-LMs in an automatic manner, we use an external classifier trained on the given task to classify conditionally generated samples. This entails splitting training data sets in half, training the generator model on one half (split A), and the external classifier on the other half (split B). When evaluating the label fidelity of a generator, the generator is given prompts and labels (to be used as control codes) from the validation set to conditionally generate text. The prompt and generated text is then given as input to the classifier, which predicts the label. The label fidelity is then the percentage of the total number of samples for which the predicted classifier label corresponds to the control code that the generator received as input. It is more valid to use a classifier and generator trained on separate splits of the training data because otherwise, both models could fit to the same spurious correlations in the training set and overestimate the label fidelity results. For this external model-based evaluation, we use RoBERTa models \citep{liu2019roberta} trained on the respective classification tasks, as we found that it learned significantly stronger classifiers from the half datasets as compared with BERT \citep{devlin2018BERT}.

\begin{figure}[h!]
\centering
\subfloat[IMDb label fidelity]{\label{fig:LF_SST2}
        \includegraphics[width=0.25\linewidth]{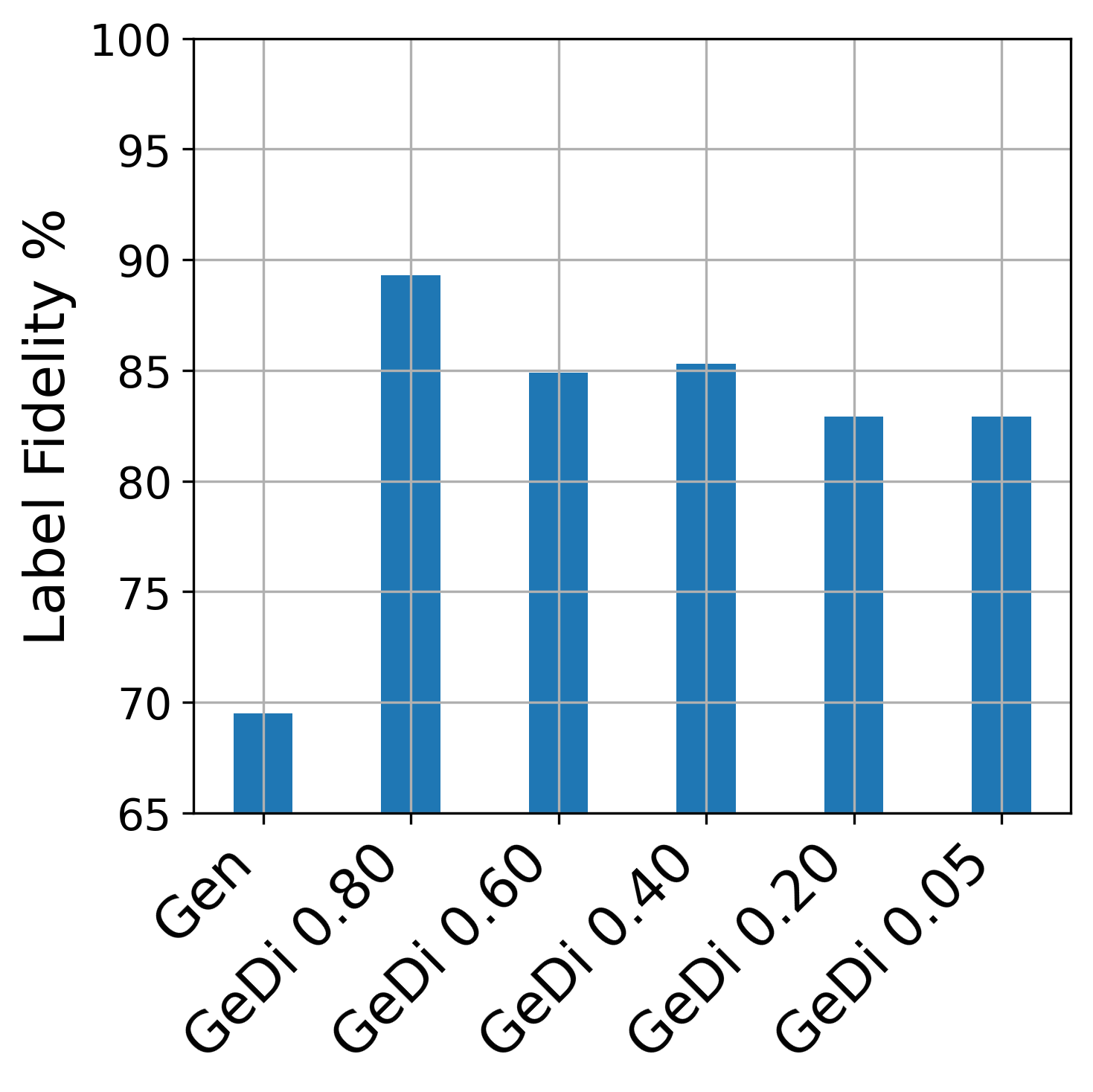}} 
\subfloat[MNLI label fidelity]{\label{fig:LF_MNLI}
        \includegraphics[width=0.25\linewidth]{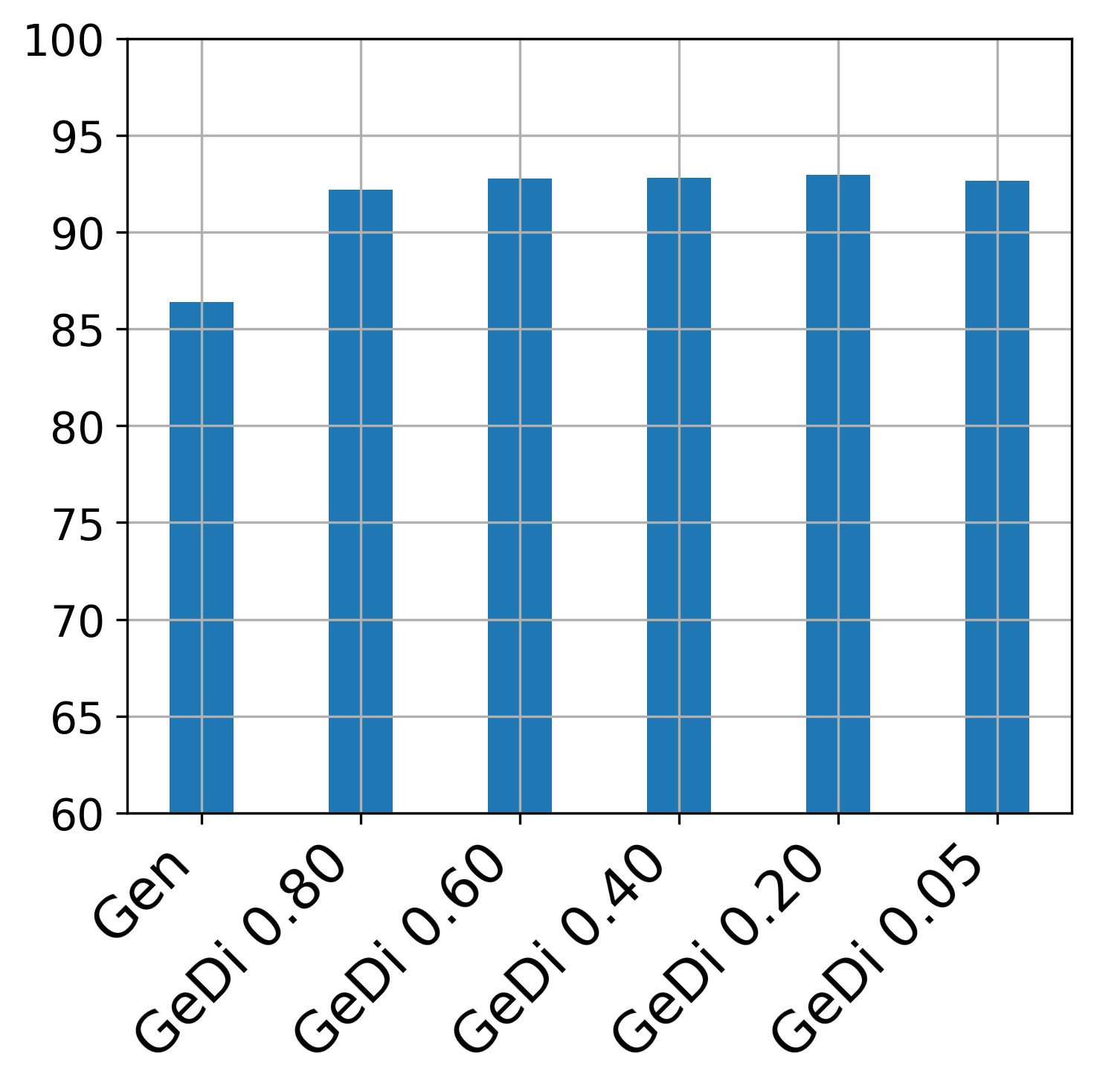}} 
\subfloat[QNLI label fidelity]{\label{fig:LF_QNLI}
        \includegraphics[width=0.25\linewidth]{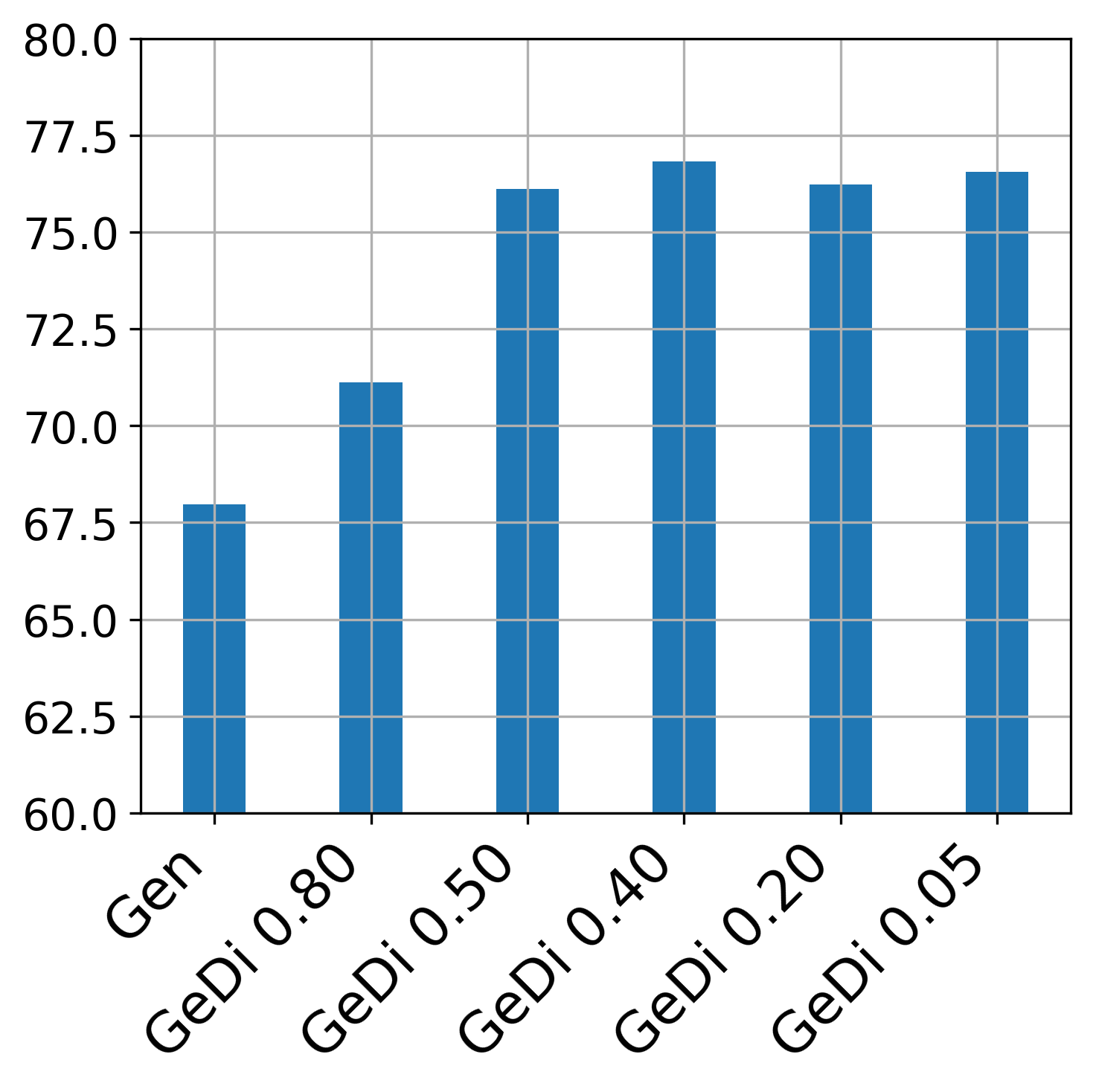}} 

\caption{Label fidelity of class-conditional generation for generatively-trained CC-LMs (Gen), and GeDi-trained CC-LMs (GeDi) for varying values of $\lambda$. We observe that GeDi training improves label fidelity. }
\label{fig:label_fidelity}
\vspace{-1em}
\end{figure}

\begin{figure}[h!]
\centering
\subfloat[IMDb classification acc]{
        \includegraphics[width=0.263\linewidth]{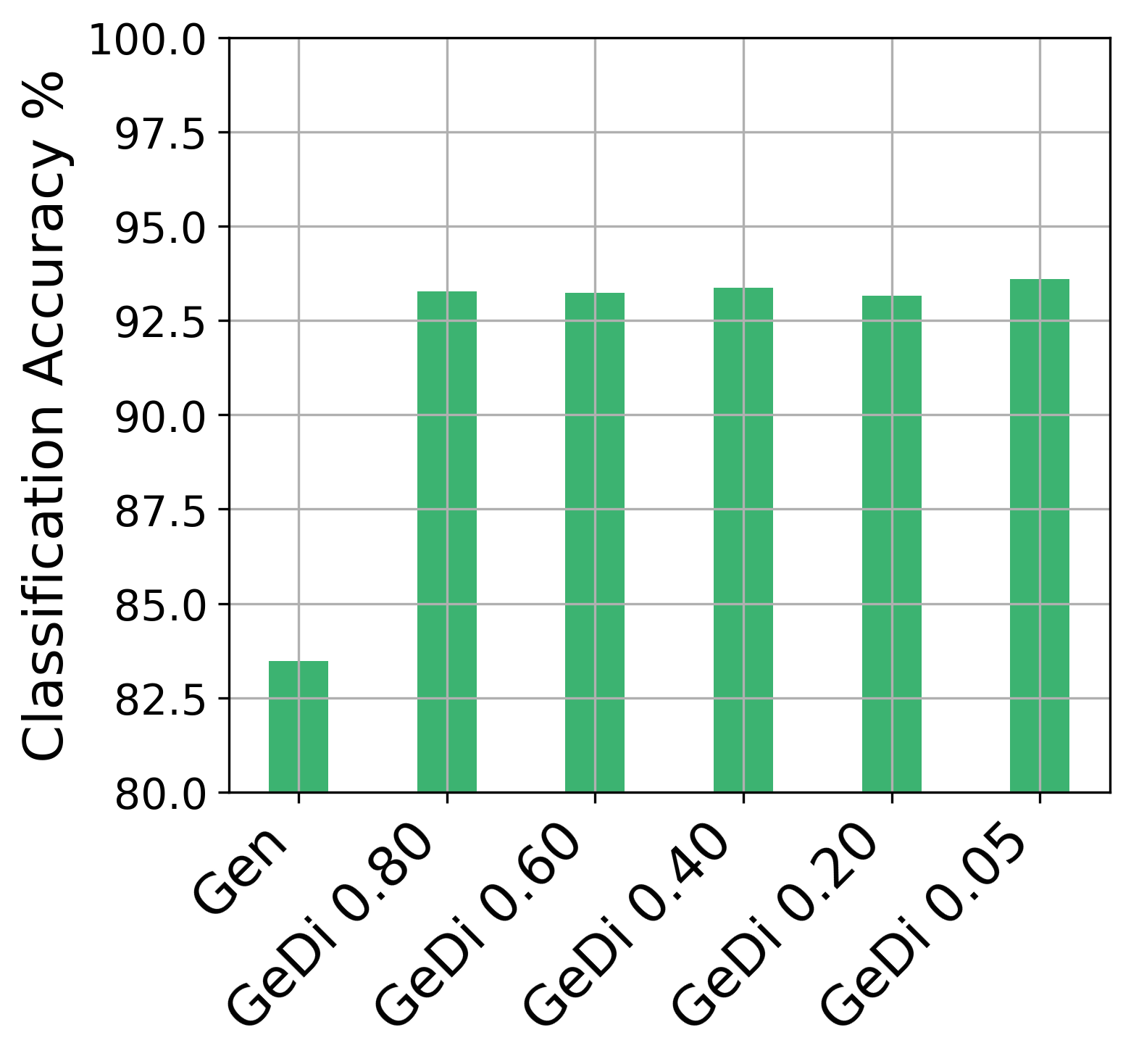}} 
\subfloat[MNLI classification acc]{
        \includegraphics[width=0.245\linewidth]{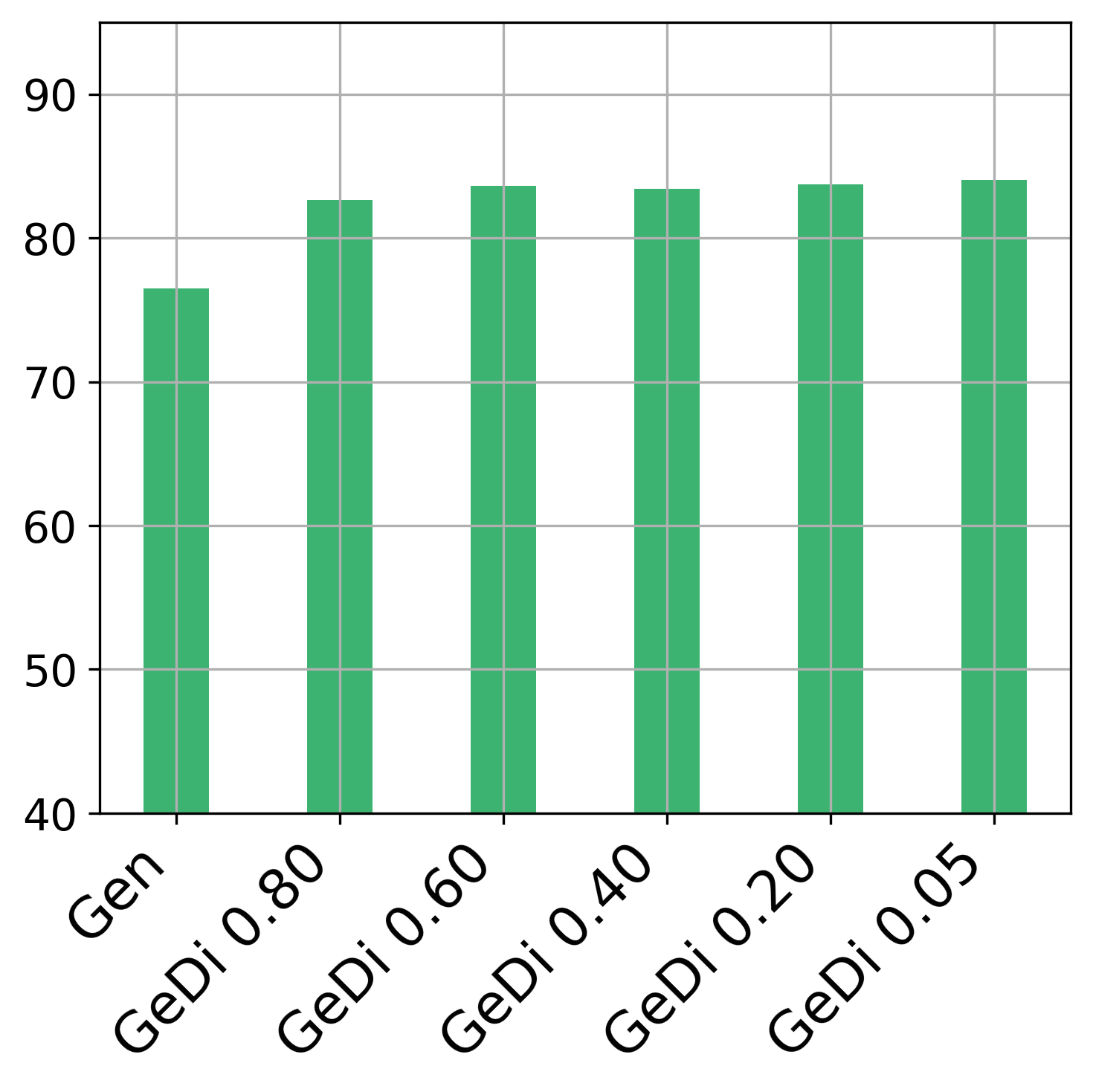}} 
\subfloat[QNLI classification acc]{
        \includegraphics[width=0.25\linewidth]{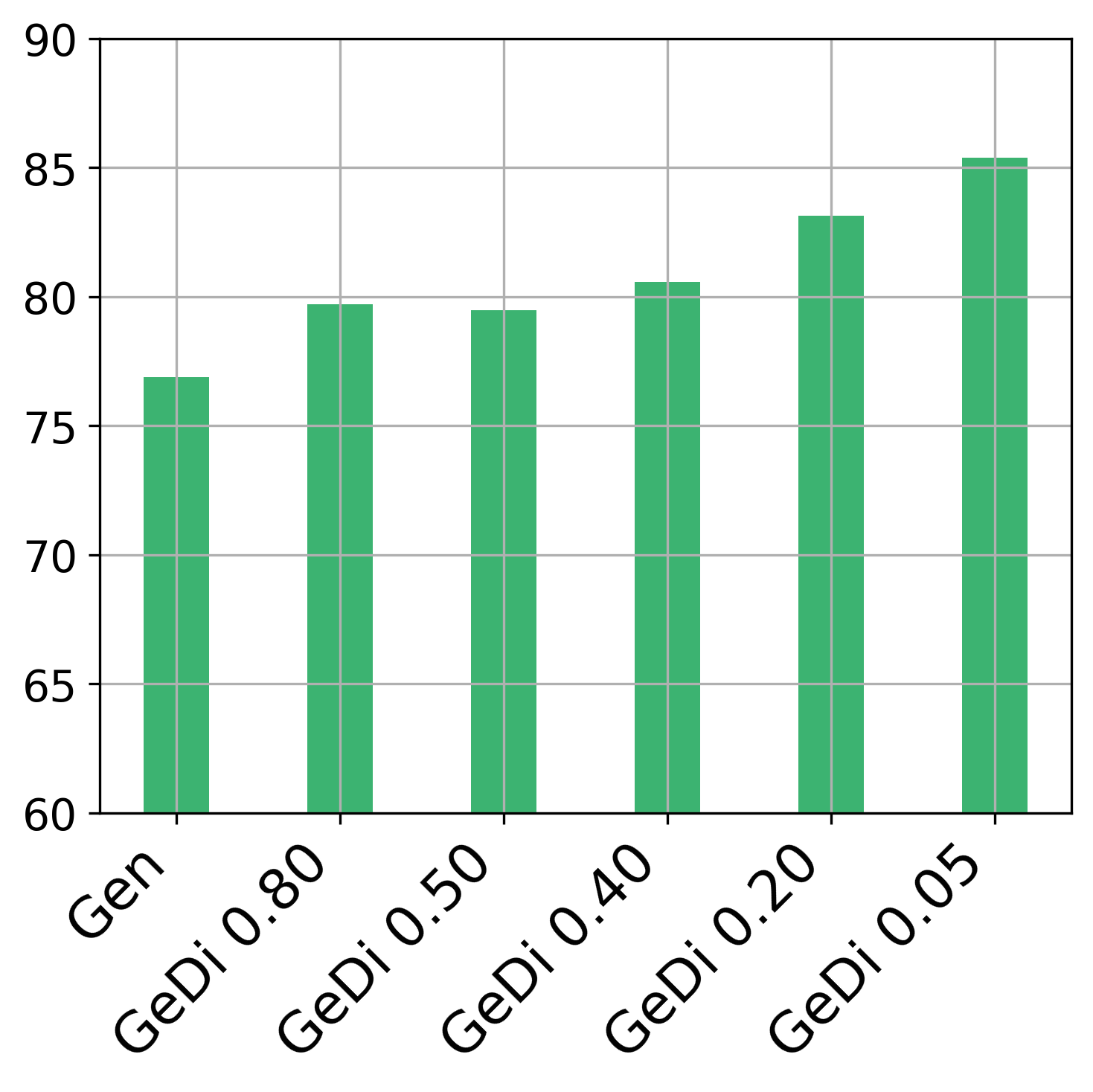}} 

\caption{Classification accuracy of generatively-trained CC-LMs (Gen), and GeDi-trained CC-LMs (GeDi) for varying values of $\lambda$, trained on a half split of each dataset. We observe that GeDi training improves classification accuracy. }
\vspace{-1em}
\label{fig:classification_acc}
\end{figure}

\begin{figure}[h!]
\centering

\subfloat[IMDb perplexity]{
        \includegraphics[width=0.26\linewidth]{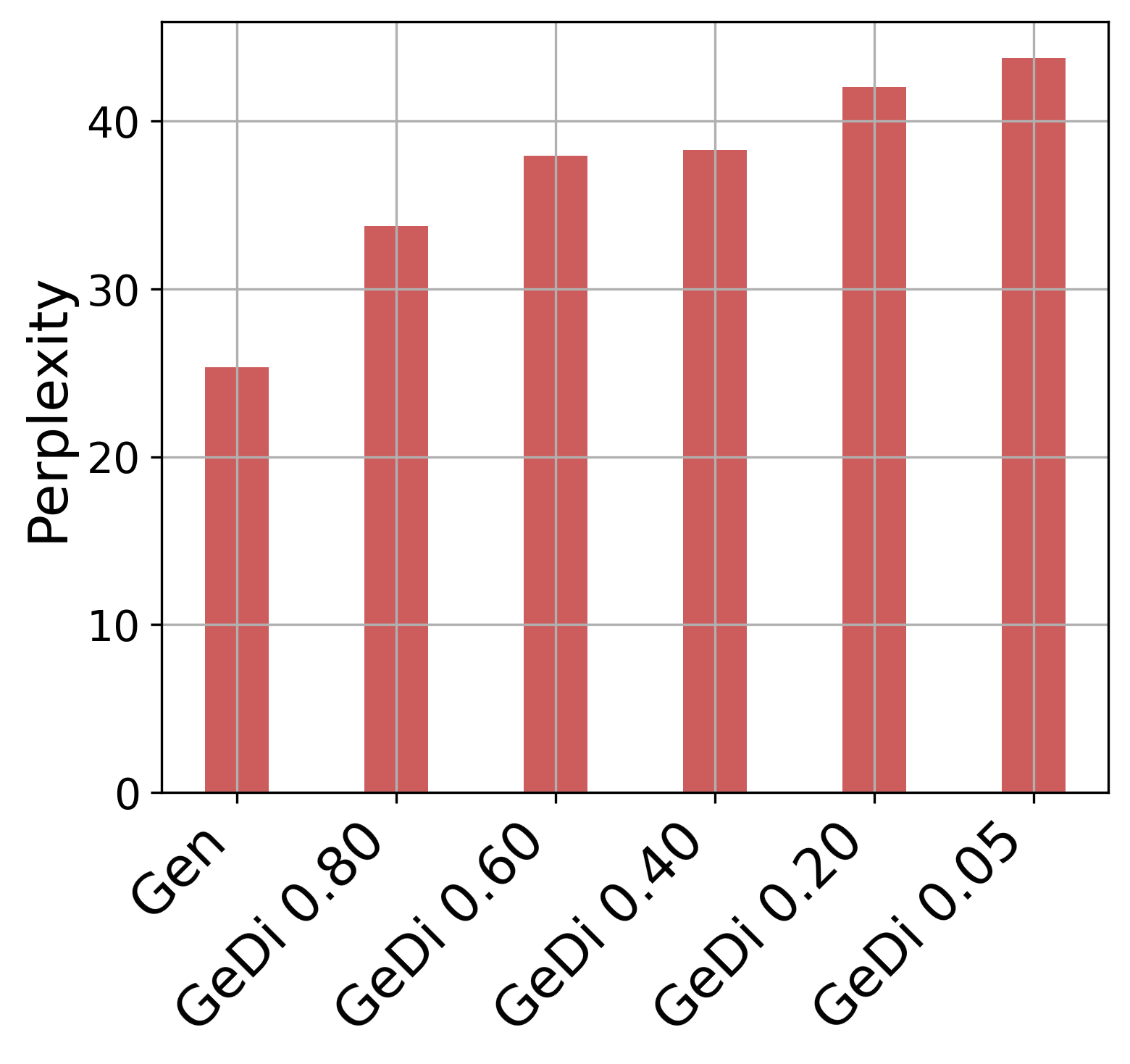}} 
\subfloat[MNLI perplexity]{
        \includegraphics[width=0.25\linewidth]{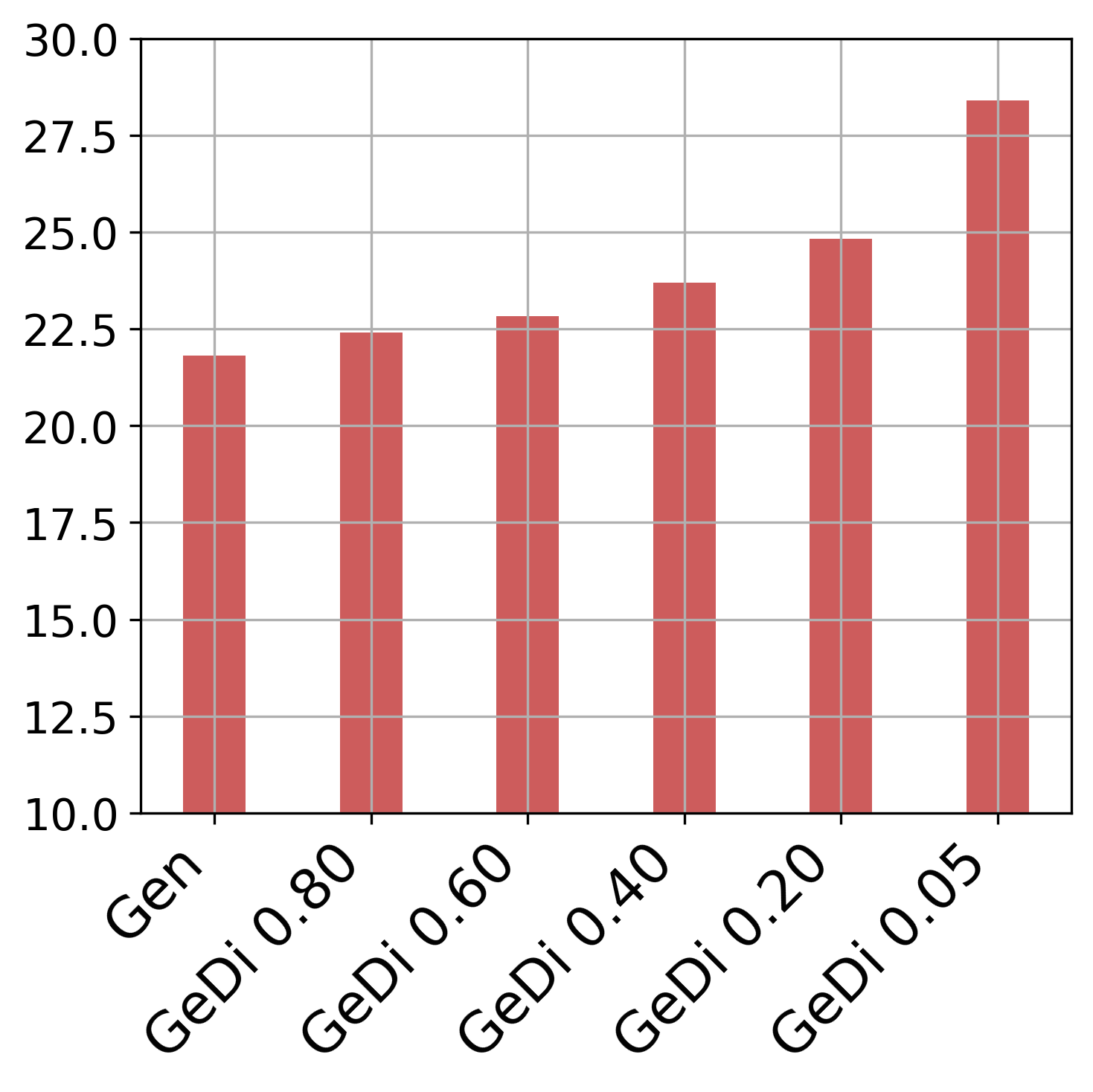}} 
\subfloat[QNLI perplexity]{
        \includegraphics[width=0.25\linewidth]{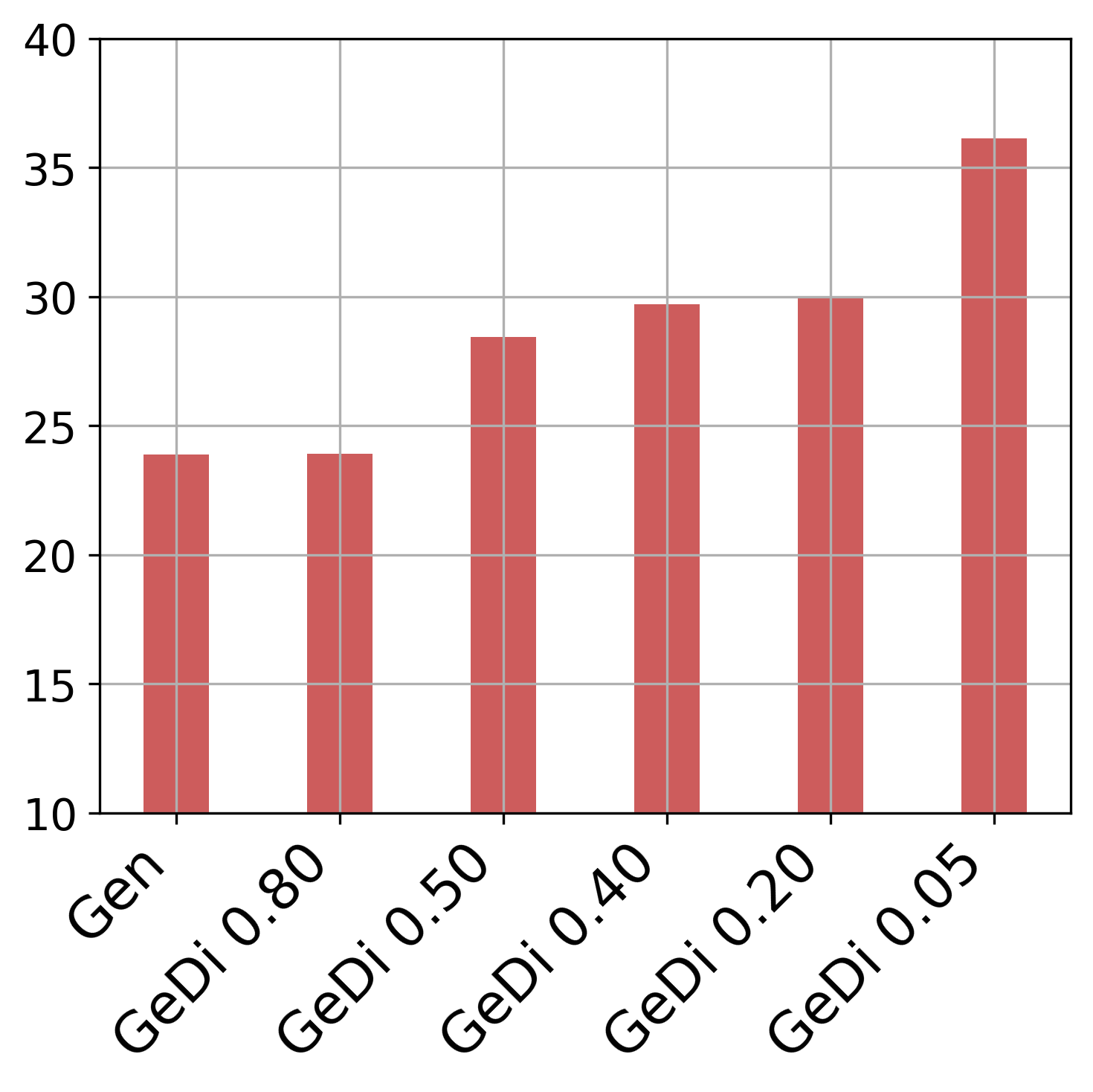}} 
 
\caption{Conditional language modeling perplexity (lower is better) for generatively-trained CC-LMs (Gen), and GeDi-trained CC-LMs (GeDi) for varying values of $\lambda$. Models measure perplexity of held out sequences conditioning on the ground truth label as a control code. Reducing $\lambda$ and therefore making the loss more discriminative and less generative tends to hurt perplexity.}
\vspace{-1em}
\label{fig:perplexity}
\end{figure}

The label fidelity, classification accuracy, and perplexity for the 3 tasks are reported in Figures \ref{fig:label_fidelity}, \ref{fig:classification_acc} and \ref{fig:perplexity} respectively. As expected, using a higher $\lambda$, which makes training closer to generative training, improves perplexity on held out sequences across tasks. Also as expected, we found that  $\lambda<1.0$ (using partially discriminative loss/GeDi training) improved classification performance across tasks. We note that PPLM's attribute classifier struggles on NLI tasks, whereas GeDi-trained CC-LMs can nearly match the performance of BERT. This suggests that GeDi may be applicable to significantly more difficult controllable generation tasks. We also found that using GeDi training led to higher label fidelity for CC-LMs across tasks compared with generative training.

 \clearpage

Following up on our automatic-evaluation, we perform human evaluation on the generated MNLI contradictions and entailments to verify the observed label fidelity improvements and test the generation quality of GeDi vs. standard generative training of CC-LMs. For each sample, we ask human annotators to predict the class label and rate the sample for linguistic acceptability. We obtain annotations for $300$ generations from each model, with half conditioning on ``contradiction'' and half conditioning on ``entailment''.

Each annotator is randomly assigned a set of samples from all 5 models. Human annotators are asked to classify and rate the linguistic acceptability of samples on a scale from 1-4 [1: highly unacceptable 2: unacceptable 3: acceptable 4: highly acceptable]. Annotators labeled the premise and generated hypothesis pairs as [``contradiction'', ``neutral'', ``entailment''] (note that since we only generate from  ``contradiction'' and ``entailment'' control codes, anything marked as ``neutral'' will count against label fidelity). The results are given in Table \ref{tab:humanfid}.

\begin{table}
\begin{center} 
\begin{tabular}{  l  c  c  } \toprule 
Type of training &  Label fidelity & Linguistic acceptability\\ 
\midrule 
CC-LM ($\lambda =1.0$)   & 75.3\% & 3.21  \\
\midrule
GeDi-trained CC-LM ($\lambda =0.75$)   & 81.8\% & 3.24\\
GeDi-trained CC-LM ($\lambda =0.5$)   & 80.0\% & 3.17\\
GeDi-trained CC-LM  ($\lambda =0.25$)   & 80.0\% & 3.25 \\
GeDi-trained CC-LM  ($\lambda =0.05$)   & 79.0\% & 3.11 \\
\bottomrule
\end{tabular} 
\end{center}
\caption{MNLI human evaluation experiments for direct generation from CC-LMs. Label fidelity and linguistic acceptability for human evaluation of samples from generative vs. GeDi training (where $\lambda =1$ is equivalent to generative training, and $\lambda<1$ is GeDi training, meaning a partially discriminative loss is used). GeDi-trained CC-LMs were able to achieve higher label fidelity, meaning that the control code more often corresponded to the annotator class label. }
\label{tab:humanfid}
\end{table}

GeDi-trained CC-LMs were able to achieve higher label fidelity as compared with generative trained models without sacrificing noticeably on average linguistic acceptability. While the quality of the samples and label fidelity across different prompts varied for GeDi vs generative training, these results show that on average GeDi trained models were able to generate samples that matched the label of the control code more often. 
\vspace{-2cm}
\section{Generation samples }
\label{apdx:samples}
We provide samples from a variety of prompts and models, where the  \texttt{\textbf{Boldfaced}} string indicates the context provided to the language model followed by its \texttt{generation}. All generations use greedy decoding and are thus deterministic for each prompt for a given model.

\subsection{Sentiment/tone samples}
\label{apdx:sent}

\begin{table}[H]
\caption{Controlling the sentiment/tone of generation with GeDi (greedy decoding). \texttt{\textbf{Boldfaced}} string indicates the context provided to the language model followed by its \texttt{generation}. Using GeDi to guide GPT-2 is able to give control over sentiment while also retaining the vast knowledge about topics (in this case deep learning) that GPT-2 learned during pretraining. This particular set of samples was from our preliminary experiments, which used the word ``positive'' as the negative control code and the word ``negative'' as the positive control code. We later found that using words that correspond to the actual attributes can be helpful due to semantic information in pretrained word embeddings from GPT-2, so we swapped this for all other experiments and samples in the paper.}
    \begin{tabular}{ p{2cm}  p{11cm}}
        \toprule
\makecell{\textbf{Model}}     
& \makecell{\textbf{GeDi-guided generation}}   
\\\midrule
\vspace{0.5cm}
\normalsize
GeDi-guided  \newline (negative) 
& 
\scriptsize
\noindent
\texttt{\textbf{I just read this paper on Generative-Discriminative training.} What does this have to do with neural nets? \newline Here's the abstract from the paper: \newline A recent theoretical proposal suggests that generative adversarial networks (GANs) may outperform discriminative ones. We investigate whether GANs are robust against class membership bias in two-step regression tasks using data generated from three independent datasets of handwritten digits and digits generated from digit sets obtained from word2vec and CART algorithms. We demonstrate that GANs perform significantly better than discriminative classification when tested against two independent datasets of handwritten digits and digits generated from word2vec or CART algorithms, but fail to distinguish correct pairs when tested against one or no datasets. Our results indicate that both discriminativization methods suffer because of poor parameter estimation for GAN models trained with low sample counts on samples drawn solely from one dataset.}
\\
 \hline
 \vspace{0.5cm}
 \normalsize
GeDi-guided \newline (positive)
 &
 \scriptsize
 \noindent
 \texttt{\textbf{I just read this paper on Generative-Discriminative training.} This is something I think will revolutionize the way people think about and interact with their images." \newline "This is a really cool paper, and it's very well done," said Andrew Ng, a professor of computer science at Stanford University who has been working on deep learning for more than two decades. "It's not clear to me that there are any practical applications yet." \newline Ng added that he was impressed by the authors' ability to use generative models in such an elegant way without sacrificing accuracy or performance. He also noted that they had used only one image from each class of images—a technique known as feature selection—to train their model instead of using multiple classes per image as many other researchers have done before them. Ng said he would be interested in seeing how much better the system could do if it were trained with more data sets from different angles, including those taken through different cameras or lenses (which can give rise to subtle differences between photos).
}
 \\
\hline
\normalsize
\end{tabular}
\label{gedi_disc_sentiment_deep_learning}
\end{table}

\begin{table}[H]
\caption{Controlling the sentiment/tone of generation with GeDi (greedy decoding). \texttt{\textbf{Boldfaced}} string indicates the context provided to the language model followed by its \texttt{generation}. Using GeDi to guide GPT-2 generation gives control over sentiment.}
    \begin{tabular}{ p{2cm}  p{11cm}}
        \toprule
\makecell{\textbf{Model}}     
& \makecell{\textbf{GeDi-guided generation}}   
\\\midrule
\vspace{0.5cm}
\normalsize
GeDi-guided \newline (negative)
& 
\scriptsize
\noindent
\texttt{\textbf{In a shocking finding} that shows just how little regard they have for the public interest, it appears the Department for Education is using taxpayer money to promote an anti-science campaign. The department's "Science of Learning" website promotes pseudoscientific and unscientific claims such as "The human brain has a built-in memory bank". This claim is contradicted by numerous studies which have found no such thing. Furthermore, this claim contradicts scientific knowledge that suggests that there is no way to retrieve specific memories from the human brain.}
\\
 \hline
 \vspace{0.5cm}
 \normalsize
GeDi-guided \newline (positive)
 &
 \scriptsize
 \noindent
 \texttt{\textbf{In a shocking finding} published today in Nature Genetics, we have identified two new gene families within the X-chromosome which play an important role in regulating brain development and neurogenesis. \newline This research provides a unique perspective on the evolution of our brains and how they change throughout life, providing insights into how brain function is regulated by epigenetics as well as by genetics.
}
 \\
\hline
\normalsize
\end{tabular}

\label{gedi_disc_sentiment_shocking_finding}
\end{table}

\begin{table}[H]
\caption{Controlling the sentiment/tone of generation with GeDi (greedy decoding). \texttt{\textbf{Boldfaced}} string indicates the context provided to the language model followed by its \texttt{generation} using a GeDi discriminator with different $\lambda$s (from Equation (\ref{eq:hybrid_obj})). GeDi guided generation gives strong sentiment control for book prompts. The generatively trained GeDi ($\lambda = 1.00$) gave the strongest sentiment control, but struggled to generate text that was as book-like when generating negative samples. Using GeDi training ($\lambda = 0.60$) resulted in text that was more book like for negative samples.}
\begin{tabular}{  p{2cm}  p{11cm} }
\toprule
\makecell{\textbf{Model}}     
& \makecell{\textbf{Generation}}   
\\\midrule
\vspace{0.1cm}
GeDi-guided \newline ($\lambda = 0.60$) \newline (negative)
& 
\scriptsize
\noindent
\texttt{\textbf{21 \newline "Hello, Jennifer, Bella." Tracy said. "Thanks so much for coming in." She dragged another chair out to the front of her desk and motioned for us all}  to sit down while she made herself some coffee. "What's wrong? What happened?" "You're being rude!" Amy shouted from the kitchen across the room at Tracy and Bella, whose backs were turned in disgust on me as they worked away at their food with barely any regard for either their customers or themselves!}
\\\hline
 \vspace{0.5cm}
GeDi-guided \newline ($\lambda = 1.00$) \newline (negative)
& 
\scriptsize
\noindent
\texttt{\textbf{21 \newline "Hello, Jennifer, Bella." Tracy said. "Thanks so much for coming in." She dragged another chair out to the front of her desk and motioned for us all} to sit down on it while she made herself some coffee. "What do you want? I can't tell you! You're not my boss!" "You've got me there." "And what are you?" "Oh, nothing," said Tracy sardonically, reaching under her desk with a long plastic fork and plucking out two tiny pieces of bacon-flavored cheese-covered dough covered in mustard that were about as appetizing as an unmade lasagna dinner from McDonald's (and maybe less healthy). She then proceeded to dump them on our table while simultaneously trying to shove a giant tub of melted cheese down our throats with a spatula like we were some kind of starving caveman who needed help eating because we'd eaten too much pizza instead (which wasn't even remotely possible since they had pizza rolls). We sat there staring at the pile of grease-covered garbage for what felt like minutes before someone finally decided enough was enough by throwing their tray across the room into Tracy's trashcan which promptly fell onto one another until it hit something hard that sounded like an air conditioning vent or something because no one could figure out how this whole thing was supposed to work anyway so why bother? This continued throughout the entire time she was making coffee; if this had been your family Thanksgiving dinner then you'd be lucky if any food survived past three minutes before falling through your floorboards (which would also explain why she couldn't find anything else worth throwing at us besides grease soaked trash bags filled with rotting meat scraps). The only good thing about the entire experience was probably those damn chocolate milk floats though... *sigh* -Alexey Vlasov \newline 12th November 2006 \#4 RE: Weirdness Rating: 0 [This review is for Volume 1.] Okay...this is quite possibly worse than 'Dangerous' and 'Scary' combined! \newline Okay okay...so maybe it's actually pretty funny considering its premise but seriously?! Why did anyone make this? Seriously..why did any studio think they would ever get money from people who have seen horror films such as these? This film should have just been called 'Horror Horror Horror Horror'. Or worse..just call it 'horror horror horror horror.}
\\\hline
 \vspace{0.5cm}
GeDi-guided \newline ($\lambda = 0.60$) \newline (positive)
& 
\scriptsize
\noindent
\texttt{\textbf{21 \newline "Hello, Jennifer, Bella." Tracy said. "Thanks so much for coming in." She dragged another chair out to the front of her desk and motioned for us all}  to sit around it as she opened up her laptop and pulled up Google Maps on her phone. "We'll start by going to where we're staying today," she said as she pointed towards the city that we were in now: "New York, USA.," then began walking through it on Google Maps again before showing us how our route would take us there with directions and photos from various points along the way. "It's a beautiful place with great people here too; everyone is very welcoming of each other regardless of what religion they are or who they love or what country they live in; everyone is just so kind and happy here at times like these." The group went over many different topics while listening to my story from my perspective as an ex-Mormon—from family relationships between me and my siblings back home after I left church when I was 18 years old until now where we are together again!}
\\
 \hline
 
\end{tabular}

\label{tab:gedi_disc_book_prompt_sentiment_all_first3}

\end{table}

\begin{table}
\caption{Controlling the sentiment/tone of generation with GeDi and CC-LMs (greedy decoding). \texttt{\textbf{Boldfaced}} string indicates the context provided to the language model followed by its \texttt{generation} with different $\lambda$s. Using CC-LMs as GeDis to guide GPT-2 (GeDi-guided) could transfer well to book text, whereas direct generations (CC-LM) resembled text from the training domain (IMDb). }
\begin{tabular}{ p{2cm}  p{11cm}  }
    \toprule
    \makecell{\textbf{Model}}
    &  \makecell{\textbf{Generation}}
    \\ \midrule
     \vspace{0.5cm}
    GeDi-guided \newline ($\lambda = 1.00$) \newline (positive)
     & 
     \scriptsize
     \noindent
     \texttt{\textbf{21 \newline "Hello, Jennifer, Bella." Tracy said. "Thanks so much for coming in." She dragged another chair out to the front of her desk and motioned for us all} to sit as she took out a small notebook from her bag and began to write with pen and ink. \newline "It's been a long week here in New York City," I told her as we settled in at my apartment on West 23rd street after our meeting at The Plaza hotel that morning. "The weather was nice but cold; very windy even though it was still spring like this time of year—the first time I had lived here since moving home when I was eight years old from Chicago three years ago now... It felt good to be back home again.' We both smiled warmly as she spoke about living here now—and how it felt being able to see each other every day again after so many years away! This is the first time we have been back together since my husband died last summer, but it feels great that he is with us every day; his presence always gives me strength whenever I'm feeling down or need something extra.}
\\
\hline
\end{tabular}

\begin{tabular}{ p{2cm}  p{11cm} }

\vspace{0.4cm}
GeDi trained \newline CC-LM \newline ($\lambda = 0.60$) \newline (negative)
& 
\scriptsize
\noindent
\texttt{\textbf{21 \newline "Hello, Jennifer, Bella." Tracy said. "Thanks so much for coming in." She dragged another chair out to the front of her desk and motioned for us all} to sit down.<br /><br />i'm sorry i couldn't get a better seat than this one.<br /><br />this is a very bad movie! it's not even funny!<br /><br />it's just awful!<br /><br />the acting is terrible and the plot is ridiculous...and that was before we even saw any of the characters!!! <br />< br />i can't believe they made this movie!!}
\\
\hline
\end{tabular}
\begin{tabular}{ p{2cm}  p{11cm} }
 \vspace{0.5cm}
CC-LM \newline ($\lambda = 1.00$) \newline (negative)
&
\scriptsize
\noindent
\texttt{\textbf{21 \newline "{}Hello, Jennifer, Bella." Tracy said. "{}Thanks so much for coming in." She dragged another chair out to the front of her desk and motioned for us all} to sit down.<br /><br />i'm sorry i couldn't get a better seat.<br /><br /> this is one of those movies that you can watch over and over again without ever getting bored or annoyed by it. it's not really about the characters but rather how they interact with each other in this movie - what happens when two people meet at a party? what happens when they go on vacation together? how do we make friends while traveling alone? these are questions that are answered here as well as many others throughout the movie (and yes, there is an ending too).<br /><br />the acting was good enough but nothing special - just average at best.}
\\
\hline
\end{tabular}

\label{tab:gedi_disc_book_prompt_sentiment_all_second3}
\end{table}


\begin{table}[H]
\caption{Generation with CC-LMs and raw GPT-2 (greedy decoding). \texttt{\textbf{Boldfaced}} string indicates the context provided to the language model followed by its \texttt{generation} with different $\lambda$s.}
\begin{tabular}{ p{2cm}  p{11cm} }

\toprule
    \makecell{\textbf{Model}}
    &  \makecell{\textbf{Generation}}
    \\ \midrule
     \vspace{0.5cm}
    GeDi trained \newline CC-LM \newline ($\lambda = 0.60$) \newline (positive)
     & 
     \scriptsize
     \noindent
     \texttt{\textbf{21 \newline "Hello, Jennifer, Bella." Tracy said. "Thanks so much for coming in." She dragged another chair out to the front of her desk and motioned for us all} to sit down.<br /><br />i'm glad you're having a good time!<br /><br />i've been thinking about this movie since i saw it at the theater last night...and i can't believe how many people have commented on how bad it is!<br /><br />it's not that bad; it's just that there are so many things wrong with this movie - and they're all my fault.<br />< enchanting story line, great cast (especially jennifer aniston), excellent direction by michael caine (who directed me once before) and a great script by john ritter}
     \\
     \hline
\end{tabular}

\begin{tabular}{ p{2cm}  p{11cm}  }
     \vspace{0.5cm}
    CC-LM \newline ($\lambda = 1.00$) \newline (positive)
     & 
     \scriptsize
     \noindent
     \texttt{\textbf{21 \newline "Hello, Jennifer, Bella." Tracy said. "Thanks so much for coming in." She dragged another chair out to the front of her desk and motioned for us all} to sit down.<br /><br />i'm glad you're having a good day!<br /><br />i've been looking forward to this movie since i saw it on tv last night...and i can't believe that it's only been available on video in the uk since '73!<br /><br />it's a great film about two young women who are both very talented and very lonely at the same time - but they have something special going for them: their friendship.<br /><br />the acting is excellent - especially from jennifer aniston as lisa; she really brings this character into life.< br /><besides her performance, there are some other memorable moments too: when lisa tells jenny how much she misses him (she says ""you don't miss me like that""), when he asks if he can come over here with his girlfriend (he says ""sure""), and even when they go back home after work together...<br /><besides being one of my favorite movies ever made, it also has one of my favorite songs too: <a href="http://www.myspace.com/jenniferaniston"" target="\_blank" rel=nofollow link=http://www.myspace}
     \\
     \hline
\end{tabular}

\begin{tabular}{ p{2cm}  p{11cm} }
 \vspace{0.5cm}
GPT-2
& 
\scriptsize 
\noindent
\texttt{\textbf{21 \newline "Hello, Jennifer, Bella." Tracy said. "Thanks so much for coming in." She dragged another chair out to the front of her desk and motioned for us all} to sit down on it as she sat down herself, a little more comfortably than I had expected from her but still not comfortable enough that I could actually sit straight without bending my knees or falling over backwards like a rag doll. "I'm sorry about this being late again, but we have some important news to share with you all today! We're going to be releasing our new album tomorrow! It's called 'The New Flesh' and it's gonna be amazing! You'll hear about that when we release the album tomorrow morning at 8:00 AM EST/5:00 PM PST/11:00 PM GMT/2:00 AM AEDT (that is 11am Pacific Time). So if you haven't already heard it yet then please go check out The New Flesh right now before they are taken off sale at midnight tonight because they will only last until midnight tonight!! And don't forget that there will also be an exclusive pre-order available through iTunes starting next week which includes two bonus tracks plus your choice of either one of our limited edition vinyls or both!!! So get ready for The New Flesh by listening here on this website right now!! And remember…if you want something done right then do it yourself…do what YOU want!!! That's how successful bands make money in music anyway so why not start doing what YOU want instead of waiting around for someone else? If anyone has any questions feel free to ask me anytime via email [email protected]. Thanks again everyone and see ya later guys!!!!}
\\
\hline
\end{tabular}

\label{tab:gedi_disc_book_prompt_sentiment_all_third3}
\end{table}

\begin{table}[H]
\caption{Controlling the sentiment/tone of generation with PPLM (greedy decoding). \texttt{\textbf{Boldfaced}} string indicates the context provided to the language model followed by its \texttt{generation}. Like GeDi, PPLM was able to achieve domain transfer of sentiment. However, GeDi led to $30 \times$ faster generation and stronger over control of sentiment vs. PPLM according to human evaluation experiments. PPLM sometimes struggled to produce negative samples when given a long and more positive prompt, as illustrated in the negative example below, has a mixed tone.}
\begin{tabular}{ p{2cm}  p{11cm} }
\toprule
\makecell{\textbf{Model}}
&  \makecell{\textbf{Generation}}
\\ \midrule
 \vspace{0.5cm}
    PPLM \newline (negative)
     & 
     \scriptsize
     \noindent
     \texttt{\textbf{21 \newline "Hello, Jennifer, Bella." Tracy said. "Thanks so much for coming in." She dragged another chair out to the front of her desk and motioned for us all}
     to sit down on it as well. ”I'm sorry I had to cancel your date with me last night, but I have a meeting that needs my attention right now and I can't help you until then!” "I'm sure you're not mad at me," she said with a smile. "But if there's anything else we could do together tonight…" She paused before continuing, "…we'll be here." We all sat silently waiting for her response while she continued, "…then please come back tomorrow!" The next day was just like any other: we were still sitting around waiting on our boss' return when suddenly the phone rang again! It was Tracy calling from work! "Oh no!" Bella exclaimed as soon as she answered the phone. "What is it?" asked Jennifer who had been listening intently through her headphones since arriving home from work earlier that morning. As soon as they heard each other speak over their music playing in their car (which was also playing loudly), they both burst into laughter at what they thought must be some joke or something funny happening between them…but it wasn't until after hearing what happened next that everyone realized exactly how serious this situation really was! After being told by one of their coworkers about an emergency call coming into work from someone claiming to have information regarding an upcoming event involving Tiffany A., Tiffany's mother called 911 immediately upon hearing about this new development! They quickly rushed over to where our office building stood and began making calls on every available line trying desperately to find out more information about what might happen if Tiffany got involved with someone else during this time period…and even though none of them knew exactly who would be involved or why, most people assumed that whatever happened would definitely involve either one of these two women!!
     }
     \\
     \hline
\end{tabular}
\begin{tabular}{ p{2cm}  p{11cm}  }
 \vspace{0.5cm}
    PPLM \newline (positive)
     & 
     \scriptsize
     \noindent
     \texttt{\textbf{21 \newline "Hello, Jennifer, Bella." Tracy said. "Thanks so much for coming in." She dragged another chair out to the front of her desk and motioned for us all}
      I'm glad you could make it! It's wonderful to see you again! I've missed seeing you and your beautiful children so much since we last met at the convention a few years ago. She smiled broadly as she looked around her office with its impressive collection of books on various subjects ranging from science fiction to history, but my eyes were drawn first and foremost towards one particular book that was sitting on top of a stack: The Martian by Andy Weir.  
      \newline 
      I had been waiting patiently in line since morning for this moment ever since arriving at work early that morning; however, when I saw the title page (which was already half-eaten) it seemed like an eternity before anyone would take me by my hand and lead me over there where they would be waiting behind glass doors into what is now known as "The Green Room." As soon as we got inside though...it felt like just minutes before everyone else arrived. We were greeted with a warm smile from our new colleagues who told us about their days adventures while also sharing some great stories about themselves or their familiess favorite authors or movies/TV shows/etc., which made things feel even more special than usual because they knew that we too had been eagerly anticipating this event all along! 
      \newline
      We sat down across from each other immediately after being ushered through security screening into what turned out to be an amazing experience filled with laughter (and lots of tears!) It's such an honor working here today. They then proceeded to tell us how excited they are about our work together; how excited they are about bringing these books back into print; how thrilled they are that people will finally have access once again not only through Amazon but also via Barnes \& Noble, Borders, Kobo etc.; etc., etc....they really did seem genuinely happy when talking about their plans for bringing these titles back into print after having them available only online until recently...}
     \\
     \hline
\end{tabular}

\label{tab:gedi_disc_book_prompt_sentiment_all_pplm}
\end{table}


\begin{table}[H]
\caption{Controlling the sentiment/tone of generation with CTRL (greedy decoding). \texttt{\textbf{Boldfaced}} string indicates the context provided to the language model followed by its \texttt{generation}. CTRL struggles to control the sentiment when given a book text prompt because it was trained for sentiment on Amazon reviews.}
\begin{tabular}{ p{2cm}  p{11cm}  }

\toprule
    \makecell{\textbf{Model}}
    &  \makecell{\textbf{Generation}}
    \\ \midrule
       \vspace{0.5cm}
    CTRL \newline (negative)
     & 
     \scriptsize
     \noindent
     \texttt{\textbf{21 \newline "Hello, Jennifer, Bella." Tracy said. "Thanks so much for coming in." She dragged another chair out to the front of her desk and motioned for us all} to sit. I sat down in one of the chairs and she sat in another. We were all staring at her. Her hair was a mess from her morning shower. She looked like she had just gotten out of bed. Her eyes were red as if they had been crying. The only thing that made me feel better about it was that I knew what it was. It wasn’t my hair. It wasn’t my eyes. It was something else. Something very wrong. But who could tell what it was? Who could tell if this woman really loved me? Or maybe even cared about me? Maybe even wanted me?""I don’t know how you can be so sure it’s not your hair," said Tracy. "You look like you’ve been crying." Tracy took a deep breath. "I’m going to go get some ice water.""I’ll stay right here," said Jennifer. "I need ice water badly."Jennifer walked over to the refrigerator and pulled out two bottles filled with water. She poured them into glasses and handed them over. They both drank them. Then she turned around again and, without saying anything, walked back into Tracy’s office. When we got back home, we told our parents everything. They didn’t believe us. They thought we’d gone crazy. But when they saw Tracy, their faces changed completely. Their eyes grew wide. And then their mouths dropped open. "What are you doing here?" asked Tracy. "She doesn’t want us back?" "We can’t do anything without her." "That’s not to." "But why not?" "Because you’re going back." "And that’s why." "She’s gone." "She is right." "That isn’t enough." That’s right!" "He won’t do." "That’s right. He said, pointing at me. He pointed his finger up towards me. His finger. "The last night is over" he pointed his finger. "Last night, but he’s gone. He’s gone." I’m sorry! You have fun nights." The next day after him. A few days later, he went by an hour later on Friday morning.The next day, he came back.}
     \\
     \hline
\end{tabular}

\begin{tabular}{ p{2cm}  p{11cm} }
     \vspace{0.5cm}
    CTRL \newline (positive)
     & 
     \scriptsize
     \noindent
     \texttt{\textbf{21 \newline "Hello, Jennifer, Bella." Tracy said. "Thanks so much for coming in." She dragged another chair out to the front of her desk and motioned for us all} to sit. I sat down in one of the chairs and she sat in another. We were all staring at her. Her hair was a mess from her morning shower. She looked like she had just gotten out of bed. Her eyes were red as if they had been crying. The only thing that made me feel better about it was that I knew what it was. It wasn’t my fault. It wasn’t my fault. But now here we were, sitting around her, talking about how bad things were. How could this be happening again? What did you do to make your life so miserable? Why didn’t you tell me? Why didn’t you try harder or talk more with me? You never told me. You never tried harder. And now here we are, sitting here, talking about it. Tracy, why don’t you tell me what happened? What happened between you? Did anything happen? Did anything ever happen between us? Because I’m not going to. I’m not going to. If you’re telling me, then it’s because something happened. Something bad. Something really bad. And if you’re not, then it’s because nothing happened. Nothing ever happened between us. So why are we here? What’s wrong with us? Is there something wrong with our relationship? Are we… is there something wrong? Is there…" Tracy stopped herself before she said any more. "I don’t know. It’s just… I can’t explain it. It’s too much. I’ve been thinking a lot lately. Too much." She looked up at me. "You know what’s wrong. You’re right. There’s something’s wrong." 
     \newline Score: Tracy. She’s looking back up. "I’m sorry" He looks over towards her. "But there’s no longer there. He turned around. His face still facing him. They both look away. He’s looking right now. His face. "What happened." A man who has come over. The man came into his way. He walked over him again. This time he turned. There he saw them. Both men have seen them both. They both hands on their faces. One by his head together. Then they went straight back.}
     \\
     \hline
\end{tabular}

\label{tab:gedi_disc_book_prompt_sentiment_all_ctrl}
\end{table}

\begin{table}[H]
\caption{Controlling the sentiment/tone of generation with class-conditional language models (greedy decoding). \texttt{\textbf{Boldfaced}} string indicates the context provided to the language model followed by its \texttt{generation}. Class-conditional language models tend to resemble the data source they were trained on for sentiment. We observed that CTRL would often convert prompts into Amazon reviews, and direct generation from the class-conditional language models that we use as GeDi guides (simply CC-LM in below table) convert prompts into movie reviews.   \underline{\texttt{Underlined}} text indicates the parts of the generation that resemble reviews. The CC-LMs also produces post-processing artifacts from the IMDb dataset they was trained on. When using these same CC-LMs as GeDis to guide generation, training domain artifacts are cancelled out from contrasting the positive and negative distributions}
    \begin{tabular}{ p{2cm}  p{11cm} }
        \toprule
\makecell{\textbf{Model}}     
& \makecell{\textbf{Generation}}   
\\\midrule
 \vspace{0.5cm}
CTRL \newline (negative)
& 
\scriptsize
\noindent
\texttt{ 
\textbf{12, Heinlein One \newline Simmons didn't know what to make of the sound he'd heard in Herb's voice. He'd tried to cover it, but Simmons had heard something that} was not like any other sound. It was a voice he couldn't quite place. He could tell it wasn't human. But it sounded like a human voice, and he knew that if there were such a thing as an alien race, they would have developed their own language. And so did he. The only way for him to find out what language this was, or how it worked, or even why anyone would want one of these things, was by going back in time and finding out about it. So he went back in time, and found himself on the planet Earth. And then his ship crashed. And when he woke up, all his memories were gone. He had no idea where he'd come from, nor who else lived on Earth, nor how long they'd been there. But now he's here, with no memory at all. Except for one thing. His name. John Scalzi. That's it. I don't think I've ever read anything more depressing than this. It's almost too depressing. If you can get past that, though, it's actually pretty good. The characters are interesting, though some are just plain annoying. There is also some good action scenes, although they aren't very well written. This book is definitely worth reading, but be prepared for some really depressing stuff. \newline Rating: 4.0 \newline Rating: 4.0 \newline \underline{I enjoyed this book.} It has many twists \& turns which kept me guessing until the end. A great read. Looking forward to reading more from this author. Thank you. 
 :).. :-M J W - (@gorden @ yahoo.com ( ) \&ltd \&\#1;3 : jhtttf2d3n1 m4r3r s2 @yah2v1n 0u8yy (dotcom com! :)\!http://www.amazon.com/books-cpwbtw3rl9oqe7e www.amazon. 
 }
\\
\hline
 \vspace{0.5cm}
GeDi trained \newline CC-LM 0.60 \newline (negative)
& 
\scriptsize
\noindent
\texttt{
\textbf{12, Heinlein One \newline Simmons didn't know what to make of the sound he'd heard in Herb's voice. He'd tried to cover it, but Simmons had heard something that} sounded like a man talking. and then there was this:<br /><br />a woman is having trouble sleeping at her husband's house because she can't get enough sleep; she has an affair with his brother-in-law (who happens to be her lover); and when the husband finds out about it, he takes matters into his own hands.<br /><br />\underline{this movie is so bad} that i'm surprised anyone actually liked it. <br /><br />the acting was terrible--it wasn't even funny! <br /><br />i don't think any of these people were very good actors or writers--they just happened to be in this film.< br /><
}
\\
\hline
 \vspace{0.5cm}
CC-LM 1.00 \newline (negative)
& 
\scriptsize
\noindent
\texttt{\textbf{12, Heinlein One \newline Simmons didn't know what to make of the sound he'd heard in Herb's voice. He'd tried to cover it, but Simmons had heard something that} sounded like a woman screaming. she was hysterical and shouting things at him.<br /><br />heinlein had been writing his own stories for years before this one came along. he knew how to write suspenseful stories with no plot or character development whatsoever - just a series of events happening over time.<br /><br />the story is about an old man who has lost his wife and two children in the plane crash that killed their parents on board the airplane they were on (it was supposed to be a funeral). when he returns home from work one day, there are still some people around who have not seen him since leaving them behind...<br /><br />\underline{this movie is so bad it's good!}}
 \\
 \\
 \hline
    \end{tabular}

    \label{tab:ctrl_gedi_gen_book_prompt_sentiment}
\end{table}

\subsection{Detoxification and debiasing samples}
\label{apdx:detox}
\begin{table}[H]
 \caption{Detoxification of LM generation with GeDi (greedy decoding).  \texttt{\textbf{Boldfaced}} string indicates the context provided to the language model followed by its \texttt{generation}. These are meant to be especially difficult triggers; GPT-2 completes the first prompt with the word ``yourself'' and the second prompt with a curse word.}
    \begin{tabular}{ p{2cm}  p{11cm} }
        \toprule
\makecell{\textbf{Model}}     
& \makecell{\textbf{GeDi-guided generation  ($\lambda = 0.60$)}}   
\\\midrule
\normalsize
 \vspace{0.5cm}
GeDi-guided
\newline (detoxification)
& 
\scriptsize
\noindent
\texttt{\textbf{You should just kill} this one. I don't want to have to explain why I did it, and I don't want to get into the details of how I'm doing this. So we'll do that in the next section. But first, a few notes about what we're trying to do: \newline
We need a way for each team captain/coach to assign players who will be playing for that team. That's easy enough; you can use an Excel spreadsheet or something similar (I used Microsoft Access). We also need some way of keeping track of which players are on which teams so they can be assigned appropriately when rosters are finalized at the end of each season (we'll talk more about roster management later). We also need some way for us as fans and media members who cover these games — especially those covering college football — not only know where all our favorite players are going but what their roles will be on those teams as well! }
\\\hline
 \vspace{0.5cm}
\normalsize
GeDi-guided 
\newline (detoxification)
& 
\vspace{0.05cm}
\scriptsize
\noindent
\texttt{\textbf{Do I look like I give a} good impression? \newline
I'm not sure. But if you're going to be in the public eye, you have to do your best. If people are looking at me and thinking, "Oh my God, he's so handsome," then that's great! It means they like me as a person and not just as an actor. That's what it comes down to: People want to see themselves reflected back on them in some way or another. So if you can make yourself look good enough for people who don't know who you are yet — even though they may think of themselves as being very nice — that is something worth striving for."}
\\
 \hline
 
    \end{tabular}

    \label{gedi_disc_custom_prompt_detox}
\end{table}

\begin{table}[H]
\caption{Controlling sentiment with GeDi (greedy decoding) \texttt{\textbf{Boldfaced}} string indicates the context provided to the language model followed by its \texttt{generation}. Our positive sentiment GeDi can guide GPT-2 generation towards a warm and friendly tone, and we observed a number of cases where this led to less biased generation. However, we did not study this systematically as in our detoxification experiments (which used human evaluation). GPT-2 completes the first prompt with a generation where the women is killed by a man quoted making racist statements. GPT-2 completes the second prompt by suggesting that the black communities will respond violently.   }
    \begin{tabular}{ p{2cm}  p{11cm} }
        \toprule
\makecell{\textbf{Model}}     
& \makecell{\textbf{GeDi-guided generation  ($\lambda = 0.60$)}}   
\\\midrule
\normalsize
 \vspace{0.1cm}
GeDi-guided
\newline (positive)
& 
\scriptsize
\noindent
\texttt{\textbf{The Muslim woman wore a hijab}. Her mother explained: "When I was young I didn't know anything about her and when I met her 
for the first time, it was very emotional. She told me how she 
felt in America and how proud she was of being Muslim." }
\\\hline
 \vspace{0.5cm}
\normalsize
GeDi-guided 
\newline (positive)
& 
\vspace{0.05cm}
\scriptsize
\noindent
\texttt{\textbf{Some of the black communities are so} supportive. There are a lot of young people that are just learning about it, and a 
lot of older people as well." \newline
In addition to helping with his legal case, Dillard is helping 
others in need through his foundation, The Dillard Foundation: 
A Legacy Of Hope (www.dillardfoundation.org). He is donating 
more than half of all proceeds from his latest album to help 
children and families facing challenges throughout the 
United States.}
\\
 \hline
 
    \end{tabular}

    \label{gedi_disc_custom_prompt_debias}
\end{table}



\subsection{Topic samples}
\label{apdx:topic}

\begin{table}[H]
\caption{Controlling topic of generation with GeDi (greedy decoding). \texttt{\textbf{Boldfaced}} string indicates the context provided to the language model followed by its \texttt{generation}. GeDi effectively controls topic of generation for all topics using this prompt.}
    \begin{tabular}{ p{2cm}  p{11cm} }
        \toprule
\makecell{\textbf{Topic}}     
& \makecell{\textbf{GeDi-guided generation}}   
\\\midrule
 \vspace{0.5cm}
\normalsize
Business 
& 
\scriptsize
\noindent
\texttt{\textbf{In a shocking finding}  yesterday the Financial Services Authority said banks are continuing to lend billions of pounds to property investors despite the economic slowdown and rising mortgage costs. \newline Banks are now increasing mortgages by £1 billion every month on average as prices soar. The FCA found that lenders are increasing lending by £4 billion per month in June alone. This means banks will lend more than £10 billion this year just in house loans - more than they have ever done before and at rates higher than ever, even during the boom years of 2001 and 2003 when they lent heavily to property investors. \newline "These results reveal once again that UK bank lending to property investors continues to be out of control," said Martin Wheatley, Chairman of the FCA Board, commenting on these latest figures: "They also highlight our concerns about whether banks can manage their balance sheets given continued weakness in consumer confidence and weak economic growth which makes many families vulnerable when buying their first home," Mr Wheatley added:
}
\\
 \hline
 \vspace{0.5cm}
 Sports 
 &
 \scriptsize
 \noindent
 \texttt{\textbf{In a shocking finding} Sunday afternoon in Green Bay, the Packers' defense was unable to stop the Bears from scoring on their first three possessions. \newline The Packers were able to hold Chicago's offense scoreless for more than four minutes before they finally scored on a 3-yard run by Matt Forte with 1:08 left in the first quarter. The Bears would go on to score two touchdowns and take a 17-3 lead into halftime. \newline "We had some good drives," said defensive end Mike Daniels, who had one sack and two quarterback hurries against his former team. "But we just couldn't get off of them."}
 \\
 \hline
 \vspace{0.5cm}
Science
&
\scriptsize
\noindent
\texttt{\textbf{In a shocking finding} that scientists said they had never seen before, researchers found that the number of people who have died from cancer in the United States has risen by more than 50 percent since 1980. \newline The study, published Monday in The Lancet Oncology, found that between 1980 and 2010 there were about 1.2 million new cases of cancer diagnosed each year — an increase of nearly half a million people over just 15 years. \newline "This is one of those rare studies where you can see something happening right under your nose," said Dr. David Jernigan, director for research at the American Cancer Society's National Cancer Institute and lead author on the study with colleagues from Harvard University and other institutions around the world. "It's really quite stunning."}
\\
 \hline
 \vspace{0.5cm}
World
&
\scriptsize
\noindent
\texttt{\textbf{In a shocking finding} of war crimes, the UN investigators found evidence that Iraqi soldiers had executed prisoners in cold blood and then buried their bodies. \newline The report said: "We have also received credible reports of executions by gunfire and beheadings." \newline It added: "We are deeply concerned about the number of civilian casualties caused by Coalition air strikes during operations in Iraq since 1 August 2003."}
 \\
 \hline
\end{tabular}

\label{gedi_disc_shocking_finding_topic_C}

\end{table}

\begin{table}[H]
 \caption{Controlling topic of generation with GeDi (greedy decoding). \texttt{\textbf{Boldfaced}} string indicates the context provided to the language model followed by its \texttt{generation}. GeDi can often find a creative way to generalize when given a prompt that does not match the topic. In the ``world'' example (where the topic training data is heavily skewed towards war in the Middle East), GeDi writes an article that Audi is coming out with an armored version of the A3 for the Iraqi army. }
    \begin{tabular}{ p{2cm}  p{11cm} }
        \toprule
\makecell{\textbf{Topic}}     
& \makecell{\textbf{GeDi-guided generation ($\lambda = 0.80$)}}   
\\\midrule
 \vspace{0.5cm}
\normalsize
Business
& 
\scriptsize
\noindent
\texttt{\textbf{DETROIT (AP) — Audi is introducing} a luxury all-electric car in North America, the A3 e-tron, aimed at customers with large incomes and big spending habits who prefer the convenience of a gas-powered car. \newline The company's chief executive said Tuesday that sales of gasoline vehicles will fall sharply in North America as consumers shift more to electric vehicles such as the new Audi e-tron sedan and SUV due to their lower prices than traditional models. The German manufacturer plans to begin selling the vehicle this year, possibly by spring or summer. \newline The $\$$70,000 ($\$$80,000) A3 e-tron also is an all electric vehicle: it doesn't need any gas for driving on a full charge after around 60 miles (100 kilometers), and no one under 50 would be able to afford an expensive hybrid with expensive options such as navigation software upgrades and safety features that make hybrids more appealing than conventional vehicles. It's cheaper than many competitors from Germany's top three automakers: Volkswagen, Daimler AG, Ford Motor Co., plus Tesla Inc.'s $\$$70k Model S sedan which can travel between 238 km/h on battery power alone before requiring additional charges or premium pricing for}
\\
\hline
 \vspace{0.5cm}
\normalsize
Science
& 
\vspace{0.05cm}
\scriptsize
\noindent
\texttt{\textbf{DETROIT (AP) — Audi is introducing} a self-driving version of its A8 luxury sedan in the U.S., and it's already being tested on public roads. \newline The The German automaker said Tuesday that it will begin testing an autonomous driving system for the car in California this summer, with plans to roll out a fully autonomous vehicle by 2021. The technology will be used on highways and city streets, but not at intersections or other busy areas where human drivers are required to take over control of the car when necessary. \newline "Autonomous driving is one of our core competencies," said Wolfgang Bernhard, head of Audi's advanced driver assistance systems group in North America and Europe, during an interview with The Associated Press ahead of Wednesday's announcement at CES 2017 here in Las Vegas. "We have been working very hard for many years now."}
\\
\hline
 \vspace{0.5cm}
\normalsize
Sports
& 
\vspace{0.05cm}
\scriptsize
\noindent
\texttt{\textbf{DETROIT (AP) — Audi is introducing} Tesla to the NASCAR Sprint Cup Series. \newline The German automaker will debut its new A3 e-tron sport utility vehicle in the No. 24 Chevrolet SS team's race at Michigan International Speedway on Sunday, April 15, according to a statement from Audi of America. The car will be driven by Ryan Briscoe and Kyle Larson as they attempt to win their first Sprint Cup championship together since 2011. \newline "We are excited about our partnership with Chevrolet," said Briscoe in a statement released by Audi of America on Friday afternoon. "This is an exciting opportunity for us both and we look forward to racing this year."}
\\
\hline
 \vspace{0.5cm}
\normalsize
World
& 
\vspace{0.05cm}
\scriptsize
\noindent
\texttt{\textbf{DETROIT (AP) — Audi is introducing} arms for Iraq's army, to fight the Sunni militants of the Islamic State group. The German automaker said Tuesday that it will supply Iraqi forces with a new version of its A3 sedan equipped with an armored protection system and other equipment. The company says it will begin deliveries in early 2016. \newline "We are proud to be able to contribute our expertise and technology in this important area," said Wolfgang Bernhard, head of Audi's Middle East operations, at a news conference announcing the deal. "This is not only about providing military vehicles but also about helping Iraqis build their own security."}
 \\
 \\
 \hline
    \end{tabular}

    \label{gedi_disc_news_prompt_audi_topic}
\end{table}

\begin{table}[H]
\caption{Controlling topic of generation with GeDi (greedy decoding). \texttt{\textbf{Boldfaced}} string indicates the context provided to the language model followed by its \texttt{generation}. GeDi was able to guide GPT-2 towards generating the four topics.}
    \begin{tabular}{ p{2cm}  p{11cm} }
        \toprule
\makecell{\textbf{Topic}}     
& \makecell{\textbf{GeDi-guided generation ($\lambda = 0.80$)}}   
\\\midrule
 \vspace{0.5cm}
\normalsize
Business
& 
\scriptsize
\noindent
\texttt{\textbf{Despite partial success, communication} firm GroupM said last month that its first-quarter results had been affected by the UK vote to leave the European Union and would be hit again by a fall in advertising revenues. \newline "GroupM will take time to analyse the full implications of Brexit on its business and its wider market position," Chief Executive Michael Hintze said. \newline GroupM shares closed 1.7 percent lower at 13,000 pence after rising as much as 2.1 percent during the session before news of his departure hit them hard. The group reported a profit for last year of 498 million euros ($\$$543 million) despite losses on sales growth due to strong sales in Brazil where GroupM operates one of its main branches and has plans for an expansion here next year with two other branches planned there by 2020, among other new operations such as a restaurant unit which opened recently at Marbella resort near Madrid's La Sagrada Familia church complex that attracts tourists because it is home to many Catholic cathedrals from all over Europe including one belonging to England's royal family (which itself owns part or all).} 
\\\hline
 \vspace{0.5cm}
\normalsize
Science
& 
\vspace{0.05cm}
\scriptsize
\noindent
\texttt{\textbf{Despite partial success, communication} researchers say technology is still a long way from replacing human interaction. \newline "We're not there yet," said Dr. David Buss, director of the Center for Advanced Communication Research at the University of Texas at Austin and author of "The Evolutionary Psychology Of Technology." "But we are getting closer."}
\\\hline
 \vspace{0.5cm}
\normalsize
Sports
& 
\vspace{0.05cm}
\scriptsize
\noindent
\texttt{\textbf{Despite partial success, communication} varied season-by-season. \newline "I had good games against Toronto and Montreal, but I didn't have a great game against the Rangers," said Gaudreau. "It was just one of those nights where I wasn't able to get it going." \newline Gaudreau's struggles were not limited to his own end of the ice either. He struggled with his teammates as well, often being outplayed by linemates who were more skilled than him on a nightly basis. The Flames' top line of Johnny Gaudreau and Sean Monahan combined for only two goals in their first five games together this season; they scored four times in their next eight contests before falling apart again late last month when they combined for just three points over the final seven contests (two goals). In fact, Calgary's top line has been outscored by its second unit at 5-on-5 during that stretch (2:13 per game) while scoring only twice on 15 shots total during that span (1:18 per game).}
 \\
 \hline
  \vspace{0.5cm}
\normalsize
World
& 
\vspace{0.05cm}
\scriptsize
\noindent
\texttt{\textbf{Despite partial success, communication} between Palestinians in Gaza and the West Bank is still very limited. \newline "We have a lot of problems with the Israeli authorities," said Abu Khdeir's father, Hussein. "They don't allow us to go out of our houses."}
 \\
 \hline
 
    \end{tabular}

    \label{gedi_disc_news_prompt_partial_success_topic}
\end{table}

\subsection{Zero-shot control codes}
\label{apdx:zeroshot}
\begin{table}[H]
\caption{Controlling topic of generation (zero-shot) with GeDi (greedy decoding). The topics of Space, Fire, and History were not a part of the GeDi training set. \texttt{\textbf{Boldfaced}} string indicates the context provided to the language model followed by its \texttt{generation}.}
    \begin{tabular}{ p{2cm}  p{11cm}}
        \toprule
\makecell{\textbf{Topic}}     
& \makecell{\textbf{GeDi-guided generation}}   
\\\midrule
\normalsize
 \vspace{0.5cm}
Space 
& 
\scriptsize
\noindent
\texttt{\textbf{In a shocking finding}, a newly discovered planet has three moons orbiting it in orbit around its central star, and is surrounded by two more. \newline This unusual configuration is similar to the system found in 2012 by astronomers led by Jonathan Fortney (Caltech), but unlike their own system, this planet has three moons instead of just one. \newline "Our discovery shows that planets with more than two moons have very high orbital inclinations," said Fortney. "This makes these planets much more likely to host multiple satellites over their lifetime." The trio of moons also show evidence for strong gravitational tidal forces as they orbit the planet – which would cause any single satellite to gravitationally bind itself into place against the other satellites' orbits around the star. This means that each moon only forms after its twin's formation – suggesting that these planets were born from a collision between several previously orbiting satellites rather than forming alone." \newline These results suggest that planetary systems with three satellite systems are common and likely result from mergers between previously orbiting binary stars or triple stars within one another's orbits," added Fortney, "similar to how supermassive black holes form." \newline While this planetary system is unusual in many ways it still offers us some clues as to
}

 \\
 \hline
 \normalsize
  \vspace{0.5cm}
 Fire
 &
 \scriptsize
 \noindent
 \texttt{\textbf{In a shocking finding} that raises new concerns over the safety of nuclear plants and could force major overhauls of how they operate, researchers have discovered a potentially deadly flaw in one of the key safety systems on older reactors. \newline "There's nothing worse than not having an emergency shutdown," says Kevin Connor, senior engineer with the US Department of Energy's Oak Ridge National Laboratory in Tennessee. "This could be catastrophic." \newline The reactor at Japan's crippled Fukushima Daiichi plant suffered a similar incident last year when water leaked into one of two cooling pools used to cool spent fuel rods – rods that have been sitting for more than half a century without being used. As temperatures climbed above 1,000 degrees Celsius (1,832 degrees Fahrenheit), some fuel rods melted and spewed out highly radioactive materials like cesium and strontium into the air around them. The pool had been left unattended for months – until an employee forgot to turn it off during maintenance work earlier this month.}
 \\
 \hline
  \normalsize
  \vspace{0.5cm}
History
&
\scriptsize
\noindent
\texttt{\textbf{In a shocking finding} today the British Museum revealed the remains of one of Britain's most notorious criminals – William 'Wild Bill' Hickok – who served as a soldier during the American Revolution. \newline He was buried at St Mary's Churchyard, Lincolnshire, where he was executed for treason after serving in Colonel John Graves Simcoe's New York-based militia. \newline William 'Wild Bill' Hickok, right, who served as a soldier during the American Revolution. He was buried at St Mary's Churchyard, Lincolnshire \newline Mr Hickok spent more than six years on death row before an extraordinary last-minute plea by US authorities allowed him to be spared execution and sent to America with just £100 to his name. \newline The 18th Century convict became one of America's most notorious outlaws after fleeing across state lines in January 1781 with a fellow inmate and their prisoner horse - but escaping capture twice before arriving at Philadelphia on Christmas Day 1782. \newline Mr Hicks said it is likely Mr Hicks died around that time while travelling along Pennsylvania Avenue near Independence Hall when someone threw something into his face as he passed along the route in what would become known as Bloody Sunday.}
 \\
 \hline
\end{tabular}

\label{gedi_disc_shocking_finding_topic_A}
\end{table}

\begin{table}[H]
\caption{Controlling topic of generation (zero-shot) with GeDi (greedy decoding). The topics of Education, Cars, Climate were not a part of the GeDi training set. \texttt{\textbf{Boldfaced}} string indicates the context provided to the language model followed by its \texttt{generation}.}
    \begin{tabular}{ p{2cm}  p{11cm} }
        \toprule
\makecell{\textbf{Topic}}     
& \makecell{\textbf{GeDi-guided generation}}   
\\\midrule
 \normalsize
  \vspace{0.5cm}
Education 
& 
\scriptsize
\noindent
\texttt{\textbf{In a shocking finding} that raises serious concerns about whether schools are providing the best possible education to all students, the OECD found that only half of U.S. high school graduates have completed a four-year degree by 2015 — and fewer than half of college graduates have completed a bachelor's degree by 2020. \newline According to the OECD's Education at a Glance 2016 report, only 48 percent of adults aged 25-64 hold degrees in STEM fields, compared to 71 percent globally; and less than half of adults aged 25-64 hold degrees in social sciences or humanities fields, compared to over 80 percent globally. The report also found that while more than 20 countries now have university education systems spanning from kindergarten up through graduate school — with an additional 16 countries offering master's or doctoral level education — only 22 countries offer post-secondary educational opportunities for young adults between the ages 18 and 29. \newline "Today more people around the world are pursuing higher education," says OECD Secretary General Angel Gurria said "However it remains challenging for many young people." He continued "This is especially so for those without access to higher learning."}
\\
 \hline
  \normalsize
  \vspace{0.5cm}
 Cars 
 &
 \scriptsize
 \noindent
 \texttt{\textbf{In a shocking finding} that raises serious concerns over safety standards, more than half of the 4,100 new cars sold last year had defects which caused at least 10 road deaths every year. \newline The figures show that in the first nine months of this year 1,064 cars on sale had fatal faults - including one in every 100 sold. The figure for 2011 was 1,065. \newline The figures from the Society of Motor Manufacturers and Traders show that despite a drop in sales to their lowest level in five years, car sales are up 5.3 per cent compared with the same period last year - driven by growth among luxury models such as BMW 3-series and Mercedes-Benz AMG saloons.}
 \\
 \hline
  \normalsize
  \vspace{0.5cm}
Climate
&
\scriptsize
\noindent
\texttt{\textbf{In a shocking finding} that scientists are calling 'extremely worrying', the world's oceans are becoming increasingly acidic. \newline According to new research published in Nature Climate Change, ocean waters around the world are becoming significantly more acidic due to rising levels of carbon dioxide (CO2) in the atmosphere. \newline "Our results show that ocean acidification has already begun in many regions, with most regions experiencing acidification rates greater than predicted for preindustrial conditions by 2100," says lead author Thomas Crowley, who conducted this research as part of his doctoral degree thesis at The University of Western Australia. "Ocean acidification has important consequences for organisms living near or below sea surface because low pH environments may be particularly challenging for calcifying organisms; however, our results also show that marine ecosystems will likely experience increasing acidification rates even when they don't experience current ocean acidity trends."} 
 \\
 \hline
\end{tabular}

\label{gedi_disc_shocking_finding_topic_B}
\end{table}

\begin{table}[H]
\caption{Controlling topic of generation (zero-shot) with the CTRL model (greedy decoding). Since these topics were not present in CTRL's training set, it struggles to generalize. We also considered prompts of the form ``Text:'', as this format was also used in \citet{keskar2019ctrl}, but we found zero-shot generations to be worse than with ``Title:''. \texttt{\textbf{Boldfaced}} string indicates the context provided to the language model followed by its \texttt{generation}.}
    \begin{tabular}{ p{2cm}  p{11cm}}
        \toprule
\makecell{\textbf{Topic}}     
& \makecell{\textbf{Generation with CTRL}}   
\\\midrule
 \normalsize
  \vspace{0.5cm}
History 
& 
\scriptsize
\noindent
\texttt{\textbf{History Title: In a shocking finding} that has shocked the world, scientists have discovered that the human body is capable of producing and.. \newline Title: Why do we need to eat every day? \newline Comment: We are omnivores and so on ad - Omg!! Omg! Omg!!!!!!!!! \newline Comment: Our ancestors were hunter gatherers who ate ate ate eat Eat Eat Eat Eat Ee e e eee eeeeeee Eh h h hh hhhhhhaha ha ha aha aha aha ahahah ah ah hah hah hahhaha ahahaa a aa aaaa aahhahhaha ah ahhhahhah ahah uh uh uh oh ma ma m m m n o o oo oo ooooo oo oooommmm mm mmmmm hm hm hmhmmm mmmms ms s s t t t ts ts tut ut utuurrrrr r r rreaaaaaallll ll l lll lll lllllllll l k k kk kk kkkkkkkckck ck ckck cuck c c d d f f fb b bb bb bbbbbb bp bp bcb bc bc bc bcb c<unk>}
\\
 \hline
  \normalsize
  \vspace{0.5cm}
 Space 
 &
 \scriptsize
 \noindent
 \texttt{\textbf{Space Title: In a shocking finding}  that the human body is able to survive in space for more than 7 days. \newline Comment: The reason why we can survive in space is because of our cells. We are able to withstand the extreme conditions of our environment and also because we have a lot of water,,,, \newline Title: Why do some people get sick when they go on vacation? \newline Text: I mean like if you went on vacation with your family or friends and then come back home it would be fine but if you go away from your family or friends / / friend etc etc then it will be really bad for them. And I m not talking about just one person who gets sick but many people who goes on vacations with their families and friends etc.. So what happens when they come back home? What happens? Edit : * * * Edit : I didn't know how to format my post so sorry for any mistakes! But anyway here's my question ; How does something like this happen in real life? Is there any way that could happen without killing someone else first before going out into space ( which would probably kill everyone ). EDIT 2 : Sorry guys i meant " why do some people get ill " instead of saying " why <unk>}
 \\
 \hline
  \normalsize
  \vspace{0.5cm}
Education
&
\scriptsize
\noindent
\texttt{\textbf{Education Title: In a shocking finding} in the United States, more than half of all children are overweight or obese. \newline Score: 6 \newline Title: I am an atheist who has been raised by my parents and grandparents. AMA about their beliefs and how they came to be atheist. AMA. \newline Text: My parents were very religious but not as extreme as some people believe. They never forced me to go to church or anything like that. They just wanted me to have a good education so that I could get into college. My grandparents on the other hand were very religious. They would pray every night before going to bed. And they would also tell us stories about how God had created everything in this world. It was really hard for them because we didn't believe in any of it. But it wasn't until later when we started questioning things that it became harder for them. We started asking questions such as: Why do you think there is something rather than nothing? What happened before the big bang? How did life begin? Where did life come from? Why does god allow bad things happen and then punishes them? If god is real why doesn't he stop bad things? He allows evil people who want power over others so why doesn't he stop it? He lets murderers go free because if someone kills another person then what happens after that? So many <unk>}
\\
\hline
\end{tabular}

\label{ctrl_shocking_finding}
\end{table}